\documentclass[12pt, onecolumn]{article}

\usepackage{geometry} 
\usepackage{datetime2} 
\usepackage{graphicx} 
\usepackage{amsmath, amssymb} 
\usepackage{hyperref} 
\usepackage{fancyhdr} 
\usepackage{lipsum} 
\usepackage{titlesec} 
\usepackage{abstract} 

\usepackage{fancyhdr}
\renewenvironment{abstract}
   {\small
    \begin{center}
    \bfseries \abstractname\vspace{-.5em}\vspace{0pt}
    \end{center}
    \list{}{
      \setlength{\leftmargin}{10mm} 
      \setlength{\rightmargin}{\leftmargin} 
    }
    \item\relax}
   {\endlist}

\titleformat{\section}
  {\normalfont\Large\bfseries}{\thesection}{1em}{}

\titleformat{\subsection}
  {\normalfont\large\bfseries}{\thesubsection}{1em}{}

\begin{document}

\title{\Large New Online Communities: Graph Deep Learning on Anonymous Voting Networks to Identify Sybils in Polycentric Governance}
\author{Quinn DuPont}

\maketitle
\begin{abstract}
This research examines the polycentric governance of digital assets in blockchain-based Decentralized Autonomous Organizations (DAOs). It offers a theoretical framework and addresses a critical challenge facing decentralized governance by developing a method to identify Sybils, or spurious identities. Sybils pose significant organizational sustainability threats to DAOs and other, commons-based online communities, and threat models are identified. The experimental method uses an autoencoder architecture and graph deep learning techniques to identify Sybil activity in a DAO governance dataset (snapshot.org). Specifically, a Graph Convolutional Neural Network (GCNN) learned voting behaviours and a fast vector clustering algorithm used high-dimensional embeddings to identify similar nodes in a graph. The results reveal that deep learning can effectively identify Sybils, reducing the voting graph by 2-5\%. This research underscores the importance of Sybil resistance in DAOs, identifies challenges and opportunities for forensics and analysis of anonymous networks, and offers a novel perspective on decentralized governance, informing future policy, regulation, and governance practices.
\end{abstract}

\section{Introduction}
Decentralized Autonomous Organizations (DAOs) represent a critical intersection of social behaviour and emerging technologies. Originating from blockchain architectures, DAOs have transcended their initial role in cryptocurrency governance to become complex social systems that defy traditional classification. In an attempt to elucidate their underlying structure and functions, this research offers a theoretical framework for DAOs and presents an inductive method to identify Sybils, thereby tackling a substantial issue facing the decentralized governance of digital assets.

A theoretical exploration of endogenous and exogenous models drawn from research on Online Communities, coupled with large scale empirical data and analysis, shapes the foundation of the neoinstitutional theory offered here. Such a theory is necessary to understand the purposes of DAOs and how they operate. In a nod to Elinor Ostrom’s work, we posit that DAOs can effectively govern Digital Common Pool Resources (DCPRs) through polycentric governance, despite their idiosyncratic challenges, if and only if they effectively manage Sybils. Sybils are fraudulent identities used to manipulate voting within digital decentralized governance (similar to `sockpuppets' or `non-player characters' in the system; the term originates in the early 2000s as part of cyber history's \emph{dramatis personae}). While traditional governance frameworks readily address the `one person, one vote' principle, the anonymity provided by blockchain technologies complicates the influence of Sybils in decentralized governance.

To address this governance challenge, we used graph deep learning techniques to identify Sybil activity on a popular DAO governance forum, snapshot.org. Specifically, our method involves a Graph Convolutional Neural Net (GCNN) with multiple engineered feature layers that learned vector node embeddings from a voting graph to construct a `similarity' subgraph, where similar nodes are identified using a fast \emph{k}-means clustering algorithm (Facebook AI Similarity Search, FAISS) and relabelled. The similarity graph is then reduced by combining predicted Sybils. While this method is transductive and sensitive to feature development and learning hyperparameters, experimental results suggest that deep learning techniques can meaningfully detect Sybils and the resulting subgraph of `true' identities is approximately 2-5\% reduced. Social scientists, policy makers, system designers, and security professionals should consider these results when studying, designing, and implementing online communities and ensure that vulnerable governance practices are made resilient to Sybil attacks.

\section{Deciphering DAOs: Advancing Research in Digital Collective Action}

The study of Decentralized Autonomous Organizations (DAOs) is marked by a burgeoning empirical landscape, rich with detailed quantitative analyses. This section focuses on the research challenges that confront DAO Studies, underscored by the nascent nature of the field, the lack of established quantitative baselines for comparison, and the complexities introduced by anonymous networks and Sybil entities.

The fluidity and lack of standardization in DAOs challenge conventional theories and frameworks of organizations, institutions, and online communities. Far removed from the hierarchical corporate models of the last century, DAOs are more like distributed, loosely coupled networks often seen in non-hierarchical collectives like social movements or even terrorist cells. A focus on their decentralized and autonomous nature, however, can be misleading; in practice, DAOs frequently fail to achieve either decentralization or autonomy in an unqualified sense.

Despite their inherent instability, variability, and quasi-anonymous infrastructure, DAOs are not beyond the scope of systematic study. In fact, their wide-ranging goals, from digital `vibes' to addressing complex real-world goals, highlights the need for scholarly attention. We have observed that increasingly new platforms are imagined, designed, built, and maintained by communities of loosely connected participants who have an economic stake in the outcome, rather than by ``core'' teams of developers with commit access, as with Bitcoin, Ethereum, and earlier centralized Web2/FLOSS platforms like Wikipedia or Linux. This makes DAOs both similar to yet distinct from existing Online Communities, a subfield of study for Information Systems.

The key unifying factor among all DAOs is their use of, or responsibility for, a decentralized ledger, or ``blockchain.'' Central to each DAO is a shared digital asset on a blockchain, often referred to as a ``token.'' Recognizing the pivotal role of ``token economics'' in these novel, community-led platforms, we find a compelling area of research: an exploration of how these digitally augmented human collectives manage shared assets. This exploration draws our attention towards the concept of a ``Common Pool Resource'' \cite{ostrom_governing_1990} and our suggested adaptation, a ``Digital Common Pool Resource'' (DCPR). The further study of DCPRs could offer valuable insights into the nature and operation of DAOs.

The importance of DAO research also goes beyond academic interest. It offers an opportunity to challenge and extend our understanding of collective action, digital resource management, and decentralized governance. Further, studying DAOs offers insights into ongoing, cutting-edge experiments attempting to create robust, long-enduring institutional forms beyond the control (and regulation) of nation states. These efforts offer an organizational buttress for new and existing cryptocurrencies (typically seeking to decentralize) and frustrate criminal investigations. More optimistically and progressively, through detailed data analysis we can better comprehend the intricacies of DAOs and tap into their potential to reshape our digital environments, perhaps even as a form of ``plural public'' \cite{jain_plural_2023}.

Before describing how the emergence of Digital Common Pool Resources (DCPRs) extends the existing neoinstitutional, resource-management perspective, we offer a brief literature review of the emerging subfield of ``DAO Studies'' by explicating existing empirical measurements and detailing the conceptual challenges these studies face. Then we briefly connect this perspective to Online Communities literature in Information Systems, in effect linking previous studies on ``Web 2.0'' to a new phenomenon in ``Web3.'' Informed by both the dynamics of online communities and commons resource management practices, we gesture towards a theory of digital asset management in a true digital commons. We conclude with an experimental analysis of a large DAO dataset (snapshot.org) and develop a graph deep learning method for detecting Sybils, addressing a key existing challenge to successful decentralized governance.

\section{Emergence and Evolution of DAO Studies}

In the early 2020s, a nascent, specialized field of ``DAO Studies'' emerged (see \cite{wang_empirical_2023}). This examination of what we now call ``Web3'' offers a new perspective on the understanding and analysis of digital collectives. While DAOs have existed formally since the emergence of the Ethereum platform in early 2015, their first year of operation was rarely studied and few, even in the industry, appreciated their eventual significance. The trajectory of this field took a definitive turn following a seminal event: the 2016 ``The DAO'' attack.

This event, which exposed vulnerabilities of DAOs but also spurred interest in and the development of decentralized governance technologies, brought about a significant shift in the discourse surrounding these structures. The 2016 The DAO attack (see \cite{dupont_experiments_2018}) serves as a key event, illuminating emerging issues surrounding the governance and legitimacy of DAOs and contributing to the theoretical foundations of a rapidly emerging DAO Studies.

Following a critical examination of this event, the field of DAO Studies started to grapple with endogenous, actor-produced theories in ways that highlighted key issues like decentralization, legitimacy, and the transformation of trust. Rooted in the experiences and perspectives of key actors in DAOs, these endogenous theories attempted to capture the often utopian vision, or social ``imaginary,'' that undergirds the diverse communities. The emergence of such a perspective, outlined by Dylan-Ennis \cite{dylan-ennis_hash_2023} and others, signifies a move towards a more nuanced understanding of the aspirations and ideologies that shape DAOs but remains locked to endogenous, actor-produced theories.

As the discourse within DAO Studies evolves, we are seeing a shift from insular, participant-derived narratives towards the integration of established socio-economic theories that offer a broader context. A pivotal theoretical framework used to analyze these phenomena has been Hirschman’s paradigm of ``Exit, Voice, and Loyalty'' \cite{hirschman_exit_1970}. In this model, `Exit' refers to the option of leaving an organization or community when its quality deteriorates or when disagreements arise, while `Voice' involves expressing dissatisfaction or working towards change from within the organization. `Loyalty,' on the other hand, reflects a commitment to the organization that might inhibit `Exit' and promote `Voice.'

In the context of DAOs, `Exit' corresponds to members' ability to disassociate from the organization, typically by selling their tokens, in response to dissatisfaction or perceived decrease in value or utility. In comparison, `Voice' manifests in the community's decision-making processes, where token holders can vote on proposals or express their opinions in online fora, thereby actively participating in shaping the organization's trajectory. The traditional notion of `Loyalty,' however, undergoes a transformation in the realm of DAOs. With the advent of `Exit to Community' (E2C) – a concept coined by Schneider and Mannan \cite{schneider_exit_2021} – `Loyalty' can morph into a mechanism for community ownership. E2C emphasizes that a loyal user or member, rather than simply exiting when dissatisfied, may have the option to help transform the organization into a more community-oriented or user-owned entity, thus introducing a novel form of `Loyalty' that redefines the relationship between community members and the DAO.

Similarly, many scholars have addressed how trust is transformed by DAOs. Much literature on the role of trust in blockchain technology presupposes that its function is to lower transaction costs within organizations, a sentiment summarized by Williamson \cite{williamson_transaction-cost_1979} with his famous theory of Transaction Cost Economics (TCE). Halaburda et al.'s \cite{halaburda_digitization_2023} research extends this argument by exploring how blockchain, as a `trust-reducing' technology, can alter the dynamics of transactional governance in organizations, including DAOs. This work explores the intricacies of how blockchain's inherent features, such as decentralization, transparency, and immutability, contribute to reducing uncertainties and transaction costs that are traditionally managed through hierarchical organizational structures. Halaburda's insights are particularly relevant in the context of DAOs, which are presupposed to operate on principles of decentralized governance and collective decision-making. By reducing the need for intermediaries and centralized control, blockchain technology enables DAOs to operate with a higher degree of trust among participants (see the `Threat Model' section \ref{threatmodel} for a description of the motivation), fostering a collaborative environment that is less susceptible to opportunistic behaviors, a key concern in TCE.

Moreover, Halaburda's perspective aligns with Hardin's theory of trust, emphasizing how trust in DAOs is not merely a function of self-interest or moral commitment but also heavily influenced by the technological framework that underpins these organizations. This viewpoint is further developed by Lemieux \cite{lemieux_searching_2022}, who reconstructs Hardin's theory \cite{hardin_tragedy_1994} to incorporate the role of blockchain technology in shaping trust dynamics. A subtle tripartite model of trust emerges from Hardin's theory, where trust is ``grounded in the truster's assessment of the intentions of the trusted with respect to some action,'' which is, typically, based in self-interest, moral commitment, and other idiosyncratic factors. Others, such as Wright and De Filippi \cite{wright_measuring_2021}, have focused on the ways technology transforms trust, drawing a parallel to the substitutability, or lack thereof, of law by code.

A central theme of these theories, and indeed within the broader field of DAO Studies, is the importance of decentralization. Motivations driving DAOs towards decentralization are everywhere. They encompass the quest for enhanced security and robustness; oftentimes there is an ideological commitment to political decentralization; and in the pockets of progressive Web3, a desire to foster a global, diverse community. As well, the discourse on decentralization naturally segues into an exploration of legitimacy and the transformation of trust, as just discussed.

However, some scholars question whether decentralization is really the right term for the communities' goals (for example, JP Vergne \cite{vergne_web3_2023} prefers ``dispersion of authority''). Others have approached the question of decentralization by measuring variables believed to be representative of an individual’s control over change. Indeed, decentralization has been a key measurement in the early quantitative and empirical literature attempting to understand DAO governance dynamics.

In practice, however, most DAOs form as part of a process where a traditional startup team has built code and close personal trust first, and then formally declare intentions to decentralize. This core team, often with external advisors, legal counsel, and early stage traditional investors, then establish an off-chain DAO for representational---not direct---democratic control. A rare few will transition to on-chain governance where the ``community'' makes decisions directly (which still need to be implemented by software developers who have commit access to open source codebases). At this point in development, decentralized governance becomes complex, particular, and institutional in nature. \hyperlink{https://docs.arbitrum.foundation/gentle-intro-dao-governance}{Arbitrum} offers a case study of this developmental path, also known as ``Progressive Decentalization'' (see \cite{dupont_progressive_2024}). The institutionalization of Arbitrum, for example, involved the launch of an off-chain DAO, a governance token launch and ``airdrop,'' a quasi-legal ``constitution,'' and a progressive decentralization of codebases led by an independent and legal foundation. 

\subsection{Quantitative and Empirical DAO Research}

The empirical landscape of DAOs has been significantly expanded by several key studies, each contributing to a deeper understanding of DAO governance and operational dynamics.

Liu et al.’s \cite{liu_illusion_2023} research provides an analysis of 50 DAOs, selected based on treasury size, number of token holders, and proposal frequency. The study’s multi-method approach, examining proposal texts from five DAOs and studying voting power, reveals a concerning voter turnout of merely 1.77\%, challenging the expected participatory norms of DAOs. This quantitative detail underpins the assertion of centralized control within DAOs, further quantified by the disproportionate voting power held by top players—35\% by the top player alone, and even more concentrated when considering the collective influence of the top three players, who control 63\% of the voting power.

Wang et al.’s \cite{wang_empirical_2023} empirical study spans a dataset of 581 DAOs with 16,246 governance proposals. Their methodology, utilizing \emph{k}-means clustering and Principal Component Analysis (PCA) on proposal titles from 2020 to 2022, brings to light the difficulty of achieving decentralization. This study’s approach, which used data scraping instead of API access, reasserts Snapshot’s claim of a 95\% coverage of DAOs, although it raises critical methodological questions, such as the accurate counting of Sybils within member counts.

Sun et al.\ \cite{sun_decentralization_2023} analyzed MakerDAO’s voting from August 5, 2019, to October 22, 2021. Their empirical analysis introduced three measurements for centralized governance: voting participation, centralized voting power via the Gini coefficient, and the largest voter’s ‘voting power.’ Their results indicate a pronounced impact of centralized governance on MakerDAO, affecting financial stability, transaction volumes, network growth, and public perception as also reflected in Twitter sentiment analyses.

Sharma et al.\ \cite{sharma_unpacking_2023} conducted a study involving ten DAOs, blending qualitative insights from interviews with quantitative on-chain data analysis. This multi-method approach found significant centralization of voting power; in particular, CompoundDAO, where out of 2,482 unique addresses, 32 ‘whale’ addresses hold between 100,000 to 10 million COMP tokens each, significantly swaying proposal outcomes.

Rikken et al.\ \cite{rikken_ins_2021} analysed 4,592 DAOs, distinguishing 3,881 with measurable activity. Their categorization is based on activity level, token holders, and deployment types, supported by a survival analysis through Receiver Operating Characteristic (ROC) curves. They find that DAOs with at least 20 token holders are likely to have a higher survival rate. The study ventures beyond mere survival indicators, engaging with emerging trends such as off-chain voting, and provocatively suggested that certain DAOs might be more accurately described as Decentralized Organizations due to these limitations.

Goldberg and Schär’s \cite{goldberg_metaverse_2023} study of Decentraland’s DAO governance analyzes 1,414 governance proposals, involving 45,333 votes from 4,345 addresses to elucidate the ‘whale effect’ and its influence on governance centralization. Their research revealed critical disparities in voting power and the strategic timing of votes, which impact the democratic integrity of the governance process.

Two further studies on MakerDAO reveal its importance to the decentralized governance of digital assets. Sun, Stasinakis, and Sermpinis \cite{sun_voter_2023} conducted a detailed investigation into voter coalitions within MakerDAO, utilizing a \emph{k}-means clustering algorithm on 809 governance proposals, uncovering the impact of voter coalitions on protocol stability and growth. Zhao et al.\ \cite{zhao_task_2022} bring an Organizational Management perspective to task management in MakerDAO, with a binary variable analysis of strategic versus operational voting among 378 proposals. Echoing others, their findings present challenges to decentralization due to the dominance of powerful stakeholders, a concern echoed across DAO studies.

Faqir-Rhazoui, Gallardo, and Hassan \cite{faqir-rhazoui_comparative_2021} present a comparative analysis of decentralized governance adoption across three main DAO platforms—Aragon, DAOstack, and DAOhaus—using data from 72,320 users across 2,353 DAO communities to evaluate growth, activity, voting systems, and funds management.

Xu et al.\ \cite{xu_autogov_2023} and Fan et al.\ \cite{fan_altruistic_2023} investigate different aspects of DAO functionality—automated decision-making and role identification, respectively. Xu et al.\ developed a governance bot using reinforcement learning for automation or decision support of DAO governance. Fan et al.\ analyzed 20,148 accounts on the popular DeFi platform Paraswap, categorizing transactional behaviours to identify participant roles and community dynamics, and highlighted the influence of speculative activities.

In an effort to understand and categorize the diversity of forms DAOs take, Wright \cite{wright_measuring_2021} proposed methodologies for measuring DAO autonomy by drawing lessons from other autonomous systems. Axelsen, Jensen, and Ross \cite{axelsen_when_2022} examined organizational typologies in DAOs to determine when a DAO can be deemed decentralized. Ziegler and Welpe \cite{ziegler_taxonomy_2022} developed a taxonomy for DAOs by exploring 35 DAOs and used agglomerative clustering to categorize them into five clusters. Finally, Braun, Haeusle, and Karpischek \cite{braun_collusion-proof_2022} attempt to model DAOs using a Transaction Cost Economics (TCE) perspective and consider the impact of revenue sharing, incentives for staking (investing), decentralization, and a variety of voting schemes. Like others who considered the role of decentralization, they model decentralization in terms of an Herfindahl-Hirschman Index (HHI), which is formally similar to Gini calculations \cite{sharma_unpacking_2023}\cite{sun_decentralization_2023} and top-player \cite{liu_illusion_2023} and whale analyses \cite{goldberg_metaverse_2023}. These calculations metaphorically determine the relative `market' position of DAOs and their `user' ranks. 

Formally, we define the following:\newline 
Herfindahl-Hirschman Index: \[HHI =\sum_{i=1}^{N} s_i^2\]
where $s_i$ is the market share of firm $i$ and $N$ is the total number of firms.\newline 
Gini: \[G = \frac{\sum_{i=1}^{N} \sum_{j=1}^{N} |x_i - x_j|}{2N^2\bar{x}}\]\newline
where \( x_i \) and \( x_j \) are the values of the income or wealth of individuals \( i \) and \( j \), \( N \) is the number of individuals, and \( \bar{x} \) is the mean income or wealth.\newline 
\( n \)-top Players: \[C_{n} = \sum_{i=1}^{n} s_i\]\newline
where \( C_n \) is the concentration of the top \( n \) players, \( s_i \) is the market share of the \( i \)-th firm, and \( n \) is the number of top firms being considered.

\begin{figure}[h]
    \centering
    \includegraphics[width=0.6\textwidth]{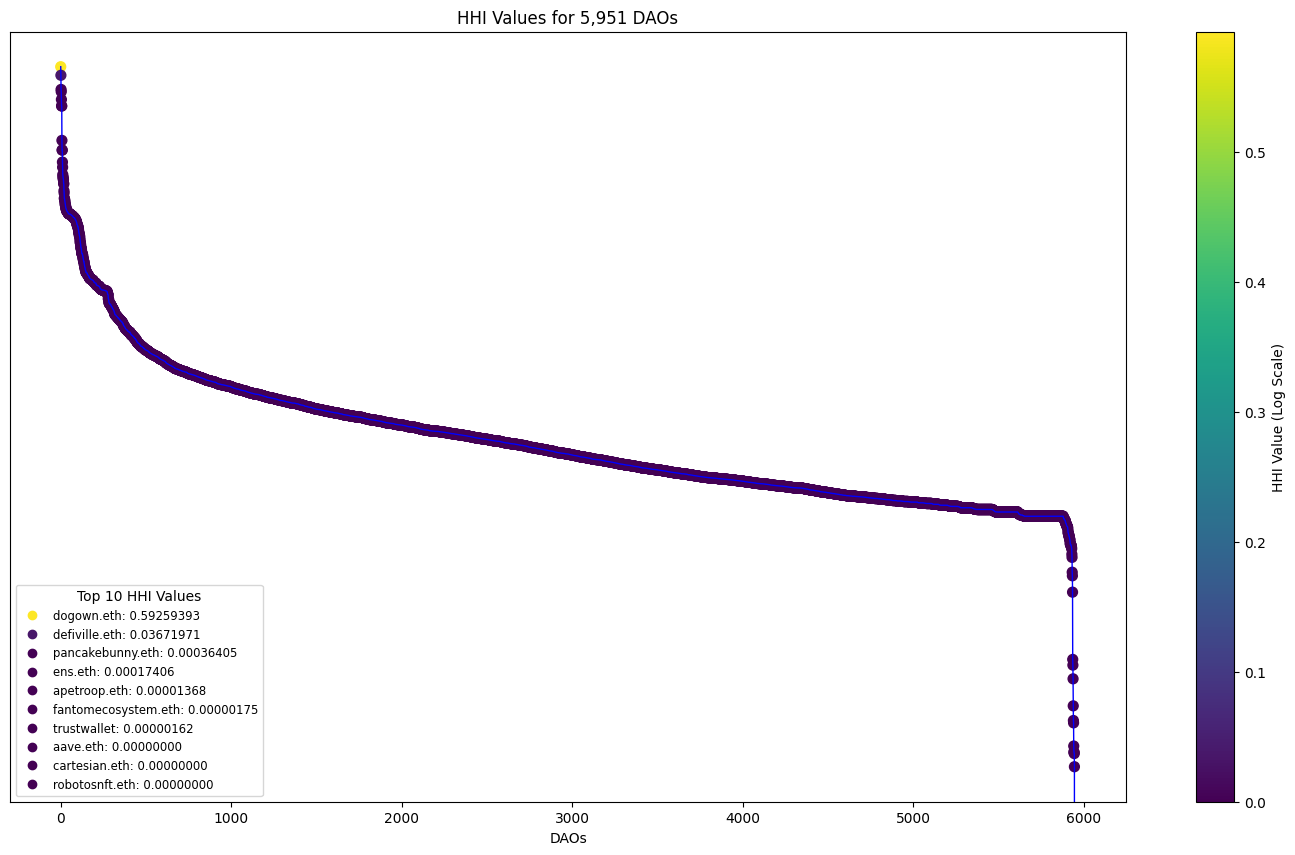}
    \caption{Herfindahl-Hirschman Index}
    \label{fig:hhi}
\end{figure}

\begin{figure}[h]
    \centering
    \includegraphics[width=0.6\textwidth]{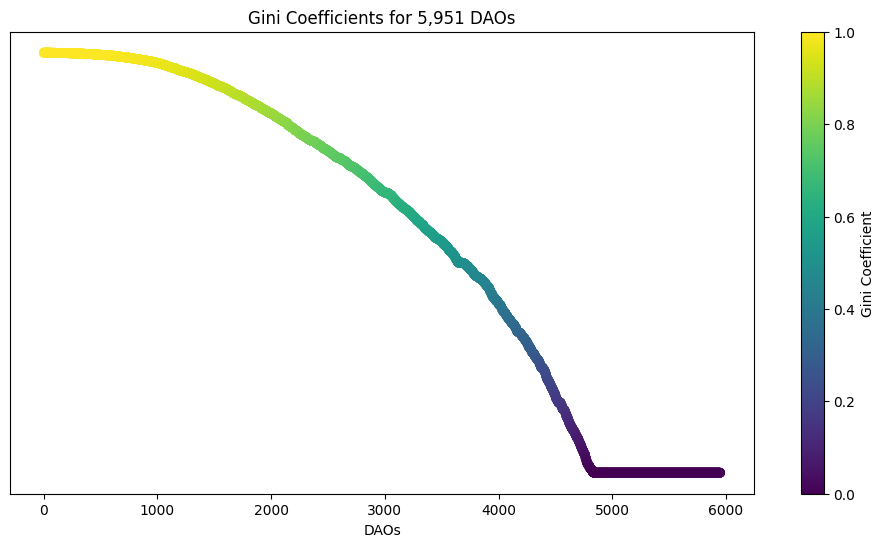}
    \caption{Gini Coefficients}
    \label{fig:gini}
\end{figure}

\begin{figure}[h]
    \centering
    \includegraphics[width=0.6\textwidth]{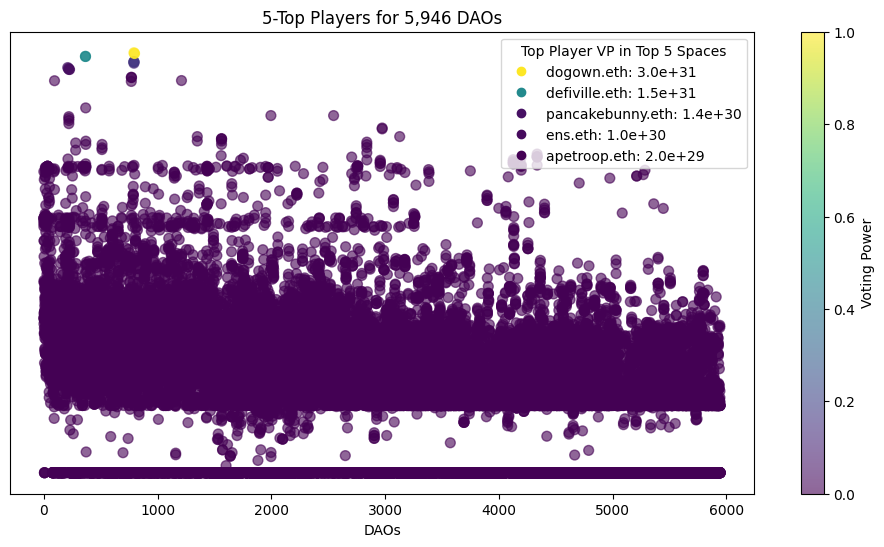}
    \caption{\emph{n}-top Players Analysis}
    \label{fig:nplayer}
\end{figure}

For comparison, we performed HHI (see Figure \ref{fig:hhi}), Gini (see Figure \ref{fig:gini}) and \( n \)-top players (see Figure \ref{fig:nplayer}) analyses on our dataset of 7.2m votes. 

Finally, Bellavitis, Fisch, and Momtaz \cite{bellavitis_rise_2023} conduct time series analyses on 2,300 DAOs using a ``DAO Analyzer'' dataset and studied voting from 1,196 DAOs using a Snapshot dataset. In the ``DAO Analyzer'' dataset they analyzed the financial returns of investments in DAO tokens, highlighting a key misunderstanding about DAOs and the impact of coin voting. While it is common practice to purchase DAO governance tokens with the expectation that they will yield financial returns like a conventional cryptocurrency, they are alsio used for governance voting. Coin voting creates a duality of motivations which muddy the conceptual waters, leading us to wonder, when do coins circulate like an economy, return value like an investment, or permit sustainable resource extraction like a commons? While the `coin' in `coin voting' functions like money, we argue below that DAOs are better understood as a commons resource management problem where coin voting determines institutional change through polycentric governance. 

In terms of governance, in the Snapshot dataset Bellavitis, Fisch, and Momtaz found that 15\% of DAOs have a token ownership threshold to make governance proposals, 160 votes on average per voter, and the proposal approval rate has hovered around 90\%. However, the voter counts are naive and do not discount the considerable noise in DAO governance nor the obvious role of Sybils, so while the results offer a helpful baseline, we question the accuracy of the findings. For instance, using slightly less naive counting methods on a much larger Snapshot dataset of 59,939 proposals, we find approximately 100 average votes per voter.

\subsection{Existing Research Challenges for DAO Studies}

DAO research is an emergent field, characterized by the absence of standardized baselines that facilitate comparative analysis. This lack of common benchmarks hinders the ability to assess the efficacy and impact of DAO governance models against established norms. Furthermore, the field is yet to develop a unified institutional theory that can frame a common set of issues and concerns for researchers to address.

The theoretical underpinnings of DAO studies are relatively undeveloped, with the initial body of literature often drawing from a `native' theory of technological utopianism that is self-referential and uncritical. This self-referential use of theory may lead to an echo chamber effect that focuses attention on decentralization and autonomy at the expense of richer social phenomena.

A significant challenge in DAO research is the anonymized nature of blockchain networks. Sybils, or duplicate identities controlled by the same entity, compound this problem, making it difficult to distinguish genuine community engagement from manipulated governance activities. The prevalence of Sybils within DAO structures can distort empirical analyses and mislead interpretations of data, resulting in flawed conclusions about the state of decentralization and participant behaviour.

These challenges necessitate the development of sophisticated methodologies and research tooling tailored to the unique characteristics of DAOs. Without addressing these methodological and theoretical gaps, the field may struggle to produce reliable and actionable insights that can inform the evolution of DAOs and their governance frameworks. Therefore, the future of DAO research lies in overcoming these hurdles to pave the way for more robust, comprehensive, and accurate studies within this dynamic and rapidly evolving domain.

\section{Online Communities}

Mindel et al.'s \cite{mindel_sustainability_2018} study provides valuable insights into the dynamics of online communities by applying the polycentric governance principles derived from Elinor Ostrom's work to Web2 online communities. The study extends the original concept of polycentric governance to digital platforms, shedding light on how decentralized systems can benefit from community-driven governance for sustainable development and resource management. Here, we want to further explore polycentric governance and align it with the challenges faced by DAOs. DAOs, as decentralized platforms, require effective management of resources and engagement without central authority. Mindel et al.'s exploration into the sustainability of polycentric information commons offers a framework that could be instrumental in addressing the new challenges facing the management of digital asset commons in Web3.

In decentralized online information systems, the principles of polycentric governance help counterbalance sustainability threats posed by unrestricted participation (which may lead to Hardin’s famous ``Tragedy of the Commons'' \cite{hardin_tragedy_1968} if left unchecked). These principles, manifested in system rules and infrastructure features, are crucial to understand how some systems decline while others flourish despite collective action threats inherent in their openness. Mindel et al.’s relevance to DAOs lies in their detailed analysis of various factors that impact the sustainability of online platforms. These include the dynamics of provision, appropriation, revitalization, and equitability, as well as collective action threats like free-riding, congestion, pollution, violation, and rebellion. By understanding these dynamics, DAOs can implement governance structures that effectively manage community contributions, maintain equitable participation, and mitigate threats to their sustainability.

\subsection{Motivation and Governance in the Transition from Web 2.0 to Web3}

Engagement in online communities has evolved from the simple pleasures of Web 2.0 to the complex incentive structures of Web3. Linus’s law, as proposed by Raymond \cite{raymond_cathedral_1999}, captured the essence of early online collaboration, suggesting that peer-driven development not only yields effective results but is also inherently enjoyable. Expanding on this, Dejean \cite{dejean_big_2015} emphasized that the allure for participants often lies in the intrinsic rewards of community engagement and recognition, over monetary gains. This perspective is increasingly relevant in the decentralized governance framework of DAOs, where traditional corporate rewards are replaced by token incentives and collective decision-making processes, as analyzed by De Filippi and Hassan \cite{filippi_blockchain_2016}.

The dichotomy of motivation in online communities is well-established, with Faraj et al. \cite{faraj_online_2016} showing that both intrinsic and extrinsic factors are critical in fostering engagement. In DAOs, smart contracts quantify and enforce commitment, potentially aligning personal motivations with communal goals. Gallus \cite{gallus_fostering_2017} indicates that symbolic rewards in communities can enhance commitment. Yet, the unique financial dynamics within DAOs complicate this simple dichotomy and underscore the need for empirical scrutiny to understand the interplay between these motivators.

Knowledge sharing and collaborative production are the lifeblood of open-source ecosystems, epitomized by Linux and Wikipedia, which stand as testaments to the power of collective intelligence. In the emergent landscape of Web3, DAOs represent a transformative expansion of these principles, infusing them with advanced decentralization and economic mechanisms. Furthermore, these structures enable novel revenue-sharing and ownership models that equitably distribute the fruits of collective efforts, challenging traditional paradigms of ownership and profit allocation. 

This evolution in the roles of contributors within DAOs is not merely a change in operations but signifies a shift in the underlying ethos of organizational structures. It heralds an era where the value generated by a community is returned to it, fostering a more engaged, motivated, and equity-driven environment. Thus, DAOs do not just adopt the foundational aspects of open-source projects; they adapt and amplify them, introducing a new model of collective action where governance and gains are intrinsically linked to participation. 

The renewed focus on social capital in online communities and DAOs becomes a cornerstone for innovation and trust. Dahlander \cite{dahlander_progressing_2011} discusses the impact of multiple affiliations on innovation, which is relevant in DAOs where decentralized control of digital assets creates new collaboration patterns. This challenges the previously held primacy of network effects, a shift underscored by Faraj et al. \cite{faraj_online_2016}.

Behavioral dynamics and social roles in online communities have been explored by researchers such as Benamar \cite{benamar_identification_2017}, but DAOs introduce a shift in these structures. Smart contracts facilitate the reimagining of roles, leading to novel governance structures within DAOs that depart from the rigidity of traditional online communities, allowing more adaptive leadership and participation models, as explored in Web 2.0 by Carillo \cite{carillo_what_2017}.

Lastly, and key to our investigation, the shift to decentralized communities presents new governance challenges. Issues such as sociability and fringe behavior, often overlooked in research, are integral to DAOs and necessitate innovative governance frameworks, a concept echoed by Faraj, Srinivas, and Wasko \cite{mcgill_university_leading_2015}. DAOs serve as a testing ground for new approaches to leadership, community engagement, and the valuation of communal resources, representing a significant departure from the governance models of traditional online communities.

\subsection{Towards Polycentric Governance}

Elinor Ostrom, in her seminal work \emph{Governing the Commons} \cite{ostrom_governing_1990}, introduced a neoinstitutional approach that challenged the notion that common pool resources (CPRs) are inevitably doomed to depletion, as suggested by the “Tragedy of the Commons” \cite{hardin_tragedy_1968}. The “Tragedy of the Commons” refers to the idea that individuals, acting in their self-interest, will deplete shared resources, leading to a loss for the entire community. This concept, first put forth by Garrett Hardin, has been the subject of extensive scholarly debate and analysis. Hardin himself later clarified that his focus was on “unmanaged commons” \cite{hardin_tragedy_1994}, and scholars such as Feeny et al.\ \cite{feeny_tragedy_1990} have explored various common-pool resource dilemmas, emphasizing the diversity of outcomes depending on governance structures.

Nonetheless, Ostrom demonstrated through empirical evidence how community-based management, rather than top-down control, could effectively govern shared resources. Her approach went beyond Hardin’s conceptualization, as she identified critical factors that enable communities to sustainably manage resources, contrary to Hardin’s more deterministic view. She challenged Hardin’s view through extensive case studies, such as fishing in Alanya, Turkey, and forest management in Nepal \cite{ostrom_governing_1990}\cite{ostrom_revisiting_1999}. Ostrom and her colleagues found that communities can and do create effective self-governance structures that prevent depletion, thereby debunking the idea that common resources are always over-exploited.

\subsubsection{Digital Common Pool Resources}

Ostrom’s ideas about polycentric governance can be adopted and adapted to digital assets and their unique governance challenges. Here, we present the novel idea of “Digital Common Pool Resources” (DCPRs) as a digital adaptation of her theory of Common Pool Resources (CPRs). In this context, resources are not physical but digital, like codebases, online identities, or cryptographic assets. Ostrom’s principles find renewed relevance in the rapidly evolving world of blockchain and cryptocurrency, where decentralized, community-based control is paramount. Scholars such as De Filippi and Hassan \cite{filippi_blockchain_2016} have further explored how blockchain technologies align with Ostrom’s principles, offering new opportunities for cooperative governance. Ostrom’s neoinstitutional approach provides a foundational framework for understanding self-governance in decentralized digital systems, focusing on the human-centered aspects of these technologies. This approach emphasizes collective action, reciprocity, and shared governance, aligning closely with the core philosophies of decentralization in the digital age.

To effectively translate Ostrom’s principles to digital assets, we must consider the nuances of decentralized governance. These principles, when reimagined for Web3, inform the design of resilient, equitable, and transparent systems. Among the challenges faced by digital commons is the cost and challenge of implementing security measures. More broadly, Ostrom’s principles provide guidance on creating sustainable digital communities:

\begin{enumerate}
    \item Clearly defined boundaries: Digital assets should have clear ownership and access rights.
    \item Proportional equivalence between benefits and costs: Participants should receive compensation proportional to their contributions.
    \item Collective-choice arrangements: Community members should have a say in setting the rules.
    \item Monitoring: The community should implement systems to monitor and validate transactions and interactions.
    \item Graduated sanctions: Penalties for rule violations should vary based on the severity and context of the breach.
    \item Conflict-resolution mechanisms: There should be accessible and fair processes for resolving disputes.
    \item Minimal recognition of rights to organize: The community’s right to self-organize should be legally recognized.
    \item Nested enterprises: Larger systems should be composed of smaller, interconnected subsystems with self-governance practices.
\end{enumerate}

By applying these principles to the decentralized governance of Web3, the creation and management of digital commons can mirror the successful self-governance of tangible resources. Our approach assumes that DAOs are responsible for DCPRs and we contribute a method for identifying Sybils to mitigate a key security challenge facing digital assets, helping ensure that these commons can be managed effectively and sustainably.

\section{Graph Deep Learning to Identify Sybils}

In decentralized governance systems, the presence of Sybils poses substantial challenges for accurately measuring user influence, decision-making, and other aspects of community structure. These spurious identities can distort the representation of user roles, creating an inaccurate picture that also makes comparisons with other Online Communities literature problematic. This research seeks to address this challenge by developing a graph deep learning method tested against a popular DAO governance forum (snapshot.org). Specifically, the method involves a Graph Convolutional Neural Net (GCNN) with multiple engineered feature layers that learn vector node embeddings from a voting graph to construct a ``similarity'' subgraph, where similar nodes are identified using a fast vector search algorithm (Facebook AI Similarity Search, aka FAISS) and relabelled into clusters. The similarity graph is then reduced by combining predicted Sybils.

\subsection{Ethical considerations}

It should be noted that using deep learning methods to analyze anonymous users on public financial and data records poses significant ethical risks and such methods should be considered a form of social network deanonymization. Even though data are ostensibly public, individuals may not have consented or even be aware that their financial transactions could be linked back to them in an identifiable way. This becomes particularly troubling given that financial data can reveal sensitive aspects of an individual's spending habits, income sources, and social networks---that they may not want publicly disclosed. In light of previous findings \cite{dupont_guiding_2020} on ethical research risks for cryptocurrencies, even when lacking real ``human subjects,'' research without consent exacerbates ethical concerns. Moreover, researchers undertaking this type of analysis must be acutely aware that their methods could be weaponized to infringe upon privacy or even perpetrate crimes.

\subsection{Sybil Resistance and Defence in Depth for Community Design}

The concept of anti-Sybil mechanisms in community design is becoming increasingly vital. In a recent podcast \cite{owocki_33_2022}, Gitcoin inventor Kevin Owocki and computer scientist Bryan Ford discuss efforts to create an identity system that counters the flaws of plutocratic governance, a particularly important issue given the widespread acceptance of coin voting in DAOs. Their recommended approach integrates principles from mechanism design and economic game theory, focusing on building identity systems that seek to preserve privacy yet provide durable digital personhood. As such, their proposed identity system would seek to provide strong anonymity and yet defend against Sybils by leveraging social ties (trust networks), centralized authorities, or maybe even other cryptographically sophisticated mechanisms (such as Zero Knowledge Proofs). 

These efforts, largely emerging from computer science, parallel the evolution of Ostrom-style commons management practices emerging in DAOs. DAOs continue to experiment with alternative identity systems, including Proof of Personhood (\hyperlink{https://whitepaper.worldcoin.org/proof-of-personhood}{PoP}), Proof of Participation (PoP), Decentralized IDs (\hyperlink{https://www.w3.org/TR/did-core/}{DIDs}), and Self Sovereign IDs (SSIDs). New identity systems that guarantee digital personhood are promising but require complex, unproven infrastructure and mass adoption. This remains an active area of research and development.

\subsubsection{Threat Model}\label{threatmodel}

\begin{figure}[h]
    \centering
    \includegraphics[width=0.6\textwidth]{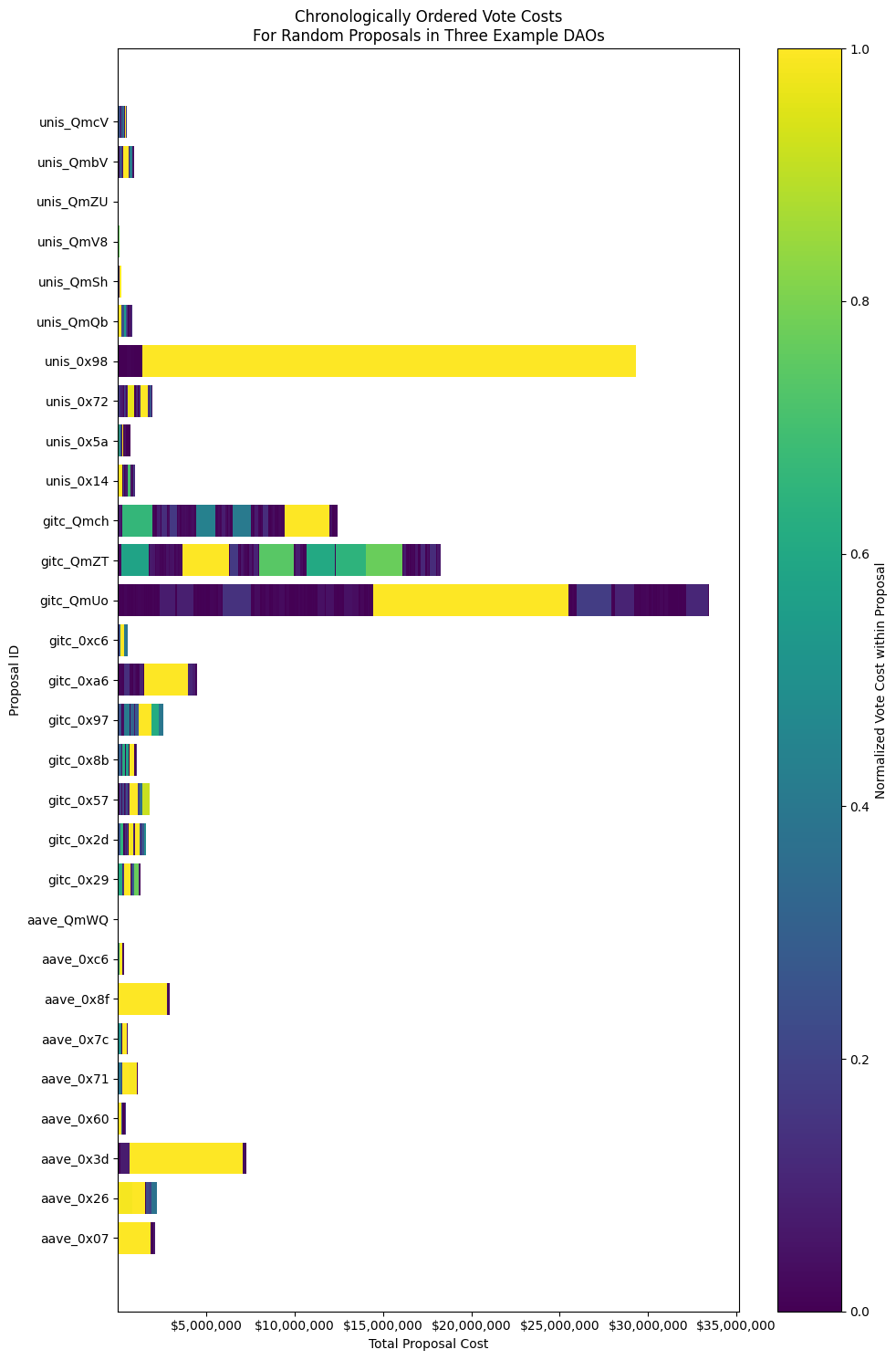}
    \caption{Minimum cost to purchase governance proposals in three example DAOs.}
    \label{fig:costofproposals}
\end{figure}

Implicitly, we have assumed that Sybils threaten governance because they can be used to mask influence, but we have yet to define the specific threat model. Our threat modeling must accept that anonymous or pseudonymous voting is seen as essential in these communities. Thus, without an identity infrastructure, an individual can easily create multiple identities to have their vote counted more than once. Since votes are tokens, an individual can create multiple identities by simply splitting their coins among multiple wallets. Given widespread economic inequality (the `whales' identified in \cite{goldberg_metaverse_2023}), `buying' a proposal is a realistic threat against DAOs (see Figure \ref{fig:costofproposals}). This attack can be multiplied by buying tokens on margin, voting, and returning the borrowed tokens---a low-cost flash attack. This overall arrangement---no durable identities and unrestricted coin voting---threatens the long-term sustainability of DAOs, which can be drained of resources, compelled to make destructive strategic decisions, or hijacked. 

Beyond plutocratic voting and community and resource misalignment, DAOs face specific threats associated with Sybils. For example, many DAOs have embraced Quadratic Voting (QV) as a way to more equitably distribute influence. Similarly, public-funding websites like Gitcoin have also embraced QV to modulate funding payouts. Specifically, if individuals 1 through \emph{n} contribute \( c_{1} \) through \( c_{n} \) to a project, then:

\[ QV = \left( \sum_{i=1}^{n} \sqrt{c_{i}} \right)^{2} \]

However, since QV awards matching funds or voting power, it is susceptible to coordinated Sybil attacks where a user may create multiple identities to extract funds or influence governance outcomes. QV is susceptible to Sybil attacks because the amplification of identities produces a squared payout, which can be used to drain resources or exert additional governance influence.

In a recent work extending the QV system, Miller, Weyl, and Erichsen \cite{miller_beyond_2022} propose an augmented form of QV that attempts to eliminate the influence of Sybils. However, the proposal is not strongly privacy preserving, since it reveals ``the intersections of social circles'' by leveraging group participation. The authors note that ``practically capturing complex social information is non-trivial'' and that the Gitcoin solution based on this work (called a ``Passport'') captures group membership information by constructing a DID record attached to the user's wallet. Users are then encouraged to submit credentials, which are added to the DID record, which is itself privacy preserving.  

Augmented QV like Gitcoin's ``Passport'' appears to offer a unique solution to a particular problem, but the proposed mechanisms are only robust against Sybils to the extent that they require attestation to group membership. Miller, Weyl, and Erichsen \cite{miller_beyond_2022} attempt to find (and formally define) appropriate ways of discounting collusion, based on identifiable group membership, and they offer several sketches for potential defence mechanisms. Three different clustering methods, a pooled centrality measurement, and an `eigen match' are proposed. However, as mentioned, the clustering methods are only effective to the extent that privacy can be reduced by exposing group membership (in theory, a user could construct a Gitcoin `Passport' using only privacy-preserving alternative identity systems, bootstrapping a DID, but this requires social behaviour and mass adoption). The proposed `offset match' is recognised as somewhat crude, since it simply pools users and discounts contributions ``to scale correlation scores so that everyone is `cared about' equally strongly.'' Finally, the `eigen match' extends the pooling mechanism by focusing on individual behaviours using an adjacency matrix to construct a set of eigenvectors. However, as I discuss below, due to the anonymous nature of blockchain systems, solutions that rely solely on edge connections are not able to address singletons, which we assume are common and prevalent. The deep learning method we propose could replace an `eigen match' component and provide more sophisticated QV Sybil defence.

\begin{figure}[ht]
    \centering
    \includegraphics[width=0.9\textwidth]{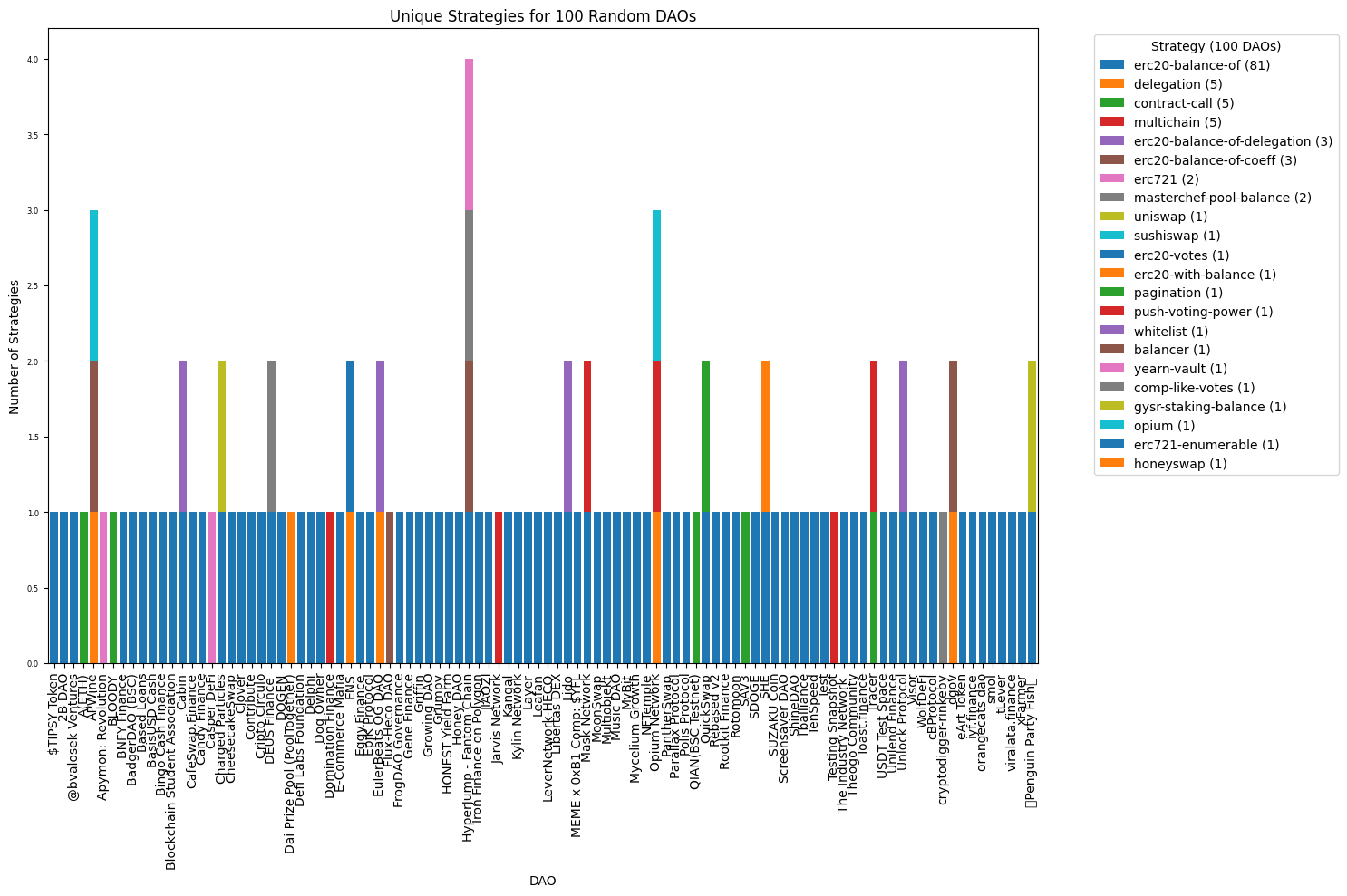}
    \caption{Unique Voting `Strategies' used by 100 DAOs.}
    \label{fig:daostrategies}
\end{figure}

On Snapshot.org, DAOs use a complex assemblages of bespoke voting ``strategies'' to ultimately calculate a user's voting power (on-chain voting like MakerDAO uses multiple smart contracts to determine voting power). For example, 81 of 100 DAOs use a simple ERC-20 token `balance' strategy (see Figure \ref{fig:daostrategies}). However, other strategies can include token delegation and arbitrary Solidity chain code execution to calculate voting power. As such, it is non-trivial to calculate the cost to `buy' a governance proposal, but with a few assumptions we can determine some baseline minimum costs. Specifically, we used the Covalent API to calculate the historical price of each ERC-20 voting token (assuming each user rationally maximized their voting power, which we know to be empirically false), and summed the costs to purchase ten random governance proposals for three example DAOs (see Figure \ref{fig:costofproposals}).

The typical free/libre and open source (FLOSS) alternative to decentralized governance has distinct challenges when tasked with sustainably and securely managing digital commons. The traditional FLOSS governance models (Linux's benevolent dictator Linus, Wikipedia's altruistic editors, Apache's Foundation model, etc.) all suffer from a reliance on personal trust or legal structuring, neither of which are robust against malicious strategic attack or free to develop in arbitrary (perhaps anti-social or illegal) ways. For the sustainability of \emph{digital commons}, the decentralization of human decision making is essential, which is why we expect to see more DAOs in the future. 

\hyperlink{https://www.getmonero.org/}{Monero}, the leading crime and privacy cryptocurrency, offers an interesting case study of the challenges facing traditional FLOSS management of valuable digital resources. Like many traditional FLOSS software projects, Monero relies on a \emph{trusted} core team of developers with commit access to a code repository. However, despite years of successful development and increasing use, Monero has recently encountered a potentially fatal organizational issue; a theft of donations (practically all cryptocurrencies encounter such threats eventually). In late 2023 a Monero wallet maintained by several core project members was drained. Accusations of internal theft pointed to a core member (@Luigi) who had recently been arrested (apparently on unrelated charges). This set of events shook up the Monero project which has led to efforts towards transition to a decentralized governance model. The long term success of Monero hinges on this transition.

In an effort to better understand the threat model facing traditionally-organized (FLOSS) cryptocurrencies like Monero, we used the GitHub API to collect all (2,984) of Monero's `Issues' and the associated comments. Each GitHub issue with its comments was then processed by a LLM (GPT-4 Turbo), prompted to identify the main issues, relevant actors, sentiment, and to categorize as either governance, security, technical, or financial. The result was further analysed by concatenating the summaries and passing these back into the LLM for further summarizing (to accommodate input token restrictions for the LLM, we vectorized the summaries, trained embeddings with a BERT pre-trained model, and clustered into appropriately sized chunks of input text). Some notable threat issues emerged from this bootstrapped history (see Appendix \ref{appendix}), including debates about infrastructure changes, community engagement and moderation, and finally, two security issues that reflect broader issues of transparency and governance. This hierarchical Retrieval Augmented Generation (RAG) method can be used at scale for rapidly generating governance histories.

Despite facing myriad governance challenges (many of which are still unknown), DAOs bring about a paradigm shift in how commons can be managed. Unlike traditional FLOSS, DAOs significantly alter expectations of transparency and trust, but demand stringent security practices. Like traditional commons, DAOs require monitoring and surveillance to prevent illegal resource extraction. But, unlike traditional commons, DAOs face a different set of challenges, especially Sybils. In response to these evolving security needs, our method of Sybil identification does not necessitate structural changes to existing governance systems and requires no new identity system. Instead, it emphasizes flexible, community-specified security responses to effectively counteract Sybil attacks. 

If Sybils can be accurately and precisely identified, communities can collectively generate rules to deal with their behaviour. As such, effective Sybil identification opens up a policy and design space (including augmenting QV). Ostrom, for instance, recommended leniency against community members who infringe the rules in traditional commons settings. Likewise, digital commons can be flexibly managed by identifying Sybils and systematically modulating voting behaviours (by attaching negative karma scores, temporarily blocking spam voting, or even using a form of social naming and shaming). Importantly, our approach is permissive to current and de facto token voting schemes and could—if perfected—permit anonymous token voting while protecting against plutocratic outcomes.

However, this flexible security response raises a crucial question: what distinguishes a spam voter from a Sybil voter, and is the distinction significant for DAOs? Conceptually, this distinction requires further analysis (is a spam voter just a Sybil from the perspective of token voting?), but from a methodological standpoint, the distinction is pressing. If we define a Sybil as a cluster of high dimensional embeddings, this implies a single hidden entity and a method for clustering based on similar vector embeddings seems appropriate. Conversely, if Sybils are equated with spam (unwanted or harmful voting), targeting anomalous node embeddings (noise) for blocking becomes more relevant. Our approach adopts the former, differentiating between spam and Sybils but more broadly it suggests a defense-in-depth strategy for DAOs, where outer layers target spam (anomalous voters) and inner layers focus on Sybil threats (clusters of similar voters).

\subsubsection{Defining Sybils and Deanonymizing Social Networks}

There is a small but varied literature on Sybils and social network deanonymization. The research often treats the problem of Sybils and the deanonymization of social networks as separate, likely because scholars tend to shape their objectives around the methodologies they have at hand. For example, Douceur's seminal work, ``The Sybil Attack'' \cite{douceur_Sybil_2002}, defines a Sybil by its possible validation—indirectly, by either a trusted, centralised name authority or a web of trust—or directly, by an identity’s ability to control networked assets. This is a strict definition of Sybils. For Sybil defence in polycentric governance contexts, however, we can adopt a much less rigid definition. For us, Sybils are clusters of \emph{distinctly} similar anonymous voters.

\begin{figure}[h]
    \centering
    \includegraphics[width=0.6\textwidth]{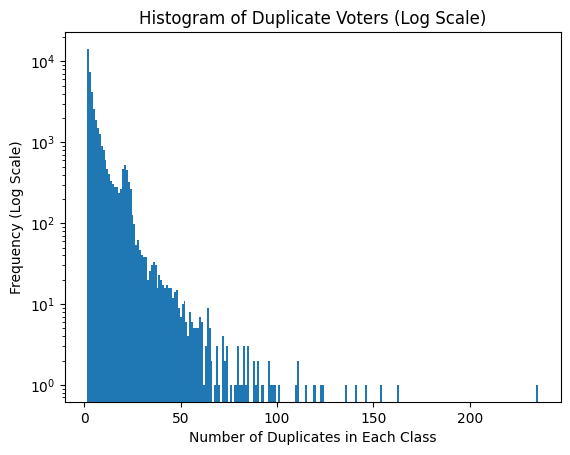}
    \caption{Histogram of Wallet Reuse in Snapshot Voting Dataset. 71\% of voters are \emph{k}$>$1.}
    \label{fig:walletreuse}
\end{figure}

\begin{figure}[h]
    \centering
    \includegraphics[width=0.6\textwidth]{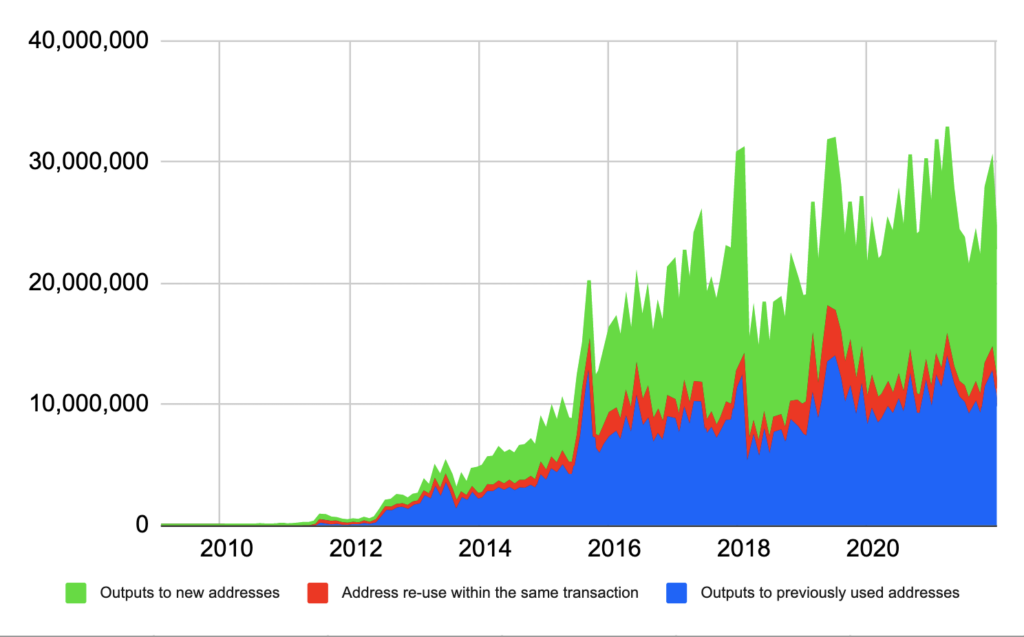}
    \caption{Bitcoin Output Count Showing Wallet Reuse from BitMex Research \cite{bitmex_research_bitcoin_2022}.}
    \label{fig:bitcoinoutputcount}
\end{figure}

We approach the challenge of identifying Sybils as a graph signal problem where we exploit poor or low \emph{k}-anonymity (wallet re-use) and use distinguishing metadata to construct higher dimensional representations to attack high \emph{k}-anonymous (``singleton'') Sybils. Sweeney’s \cite{sweeney_k-anonymity_2002} foundational theory of \emph{k}-anonymity was initially designed to protect privacy by ensuring that any released data could not be traced back to less than \emph{k} individuals. In the simplest terms, the theory is about blending in to stay anonymous. Some of our dataset is \emph{k}=1 (singletons), however 71\% of unknown users reuse wallet addresses (see Figure \ref{fig:walletreuse}), which we capture in our graph signal as multi-edge connectivity (our target users are `unknown' in the sense of having no Ethereum Name Service registration). These are the easier cases and despite a significant impact on privacy, most users on most chains reuse wallets (e.g., in Bitcoin, wallet reuse has hovered around 50-75\%; see Figure \ref{fig:bitcoinoutputcount} from \cite{bitmex_research_bitcoin_2022}). As explained below, our method to overcome \emph{k}-anonymous datasets is deceptively simple: we generate node vector embeddings and cluster the high dimensional embeddings to identify similar, unknown voters. Importantly, since we generate (higher dimensional) node embeddings for all nodes with features, we bypass the protections of \emph{k}-anonymity, making it possible to spot Sybils more effectively. 

Our method is far from novel, but it considers new contexts and uses sophisticated deanonymization techniques only recently made possible with advances in machine learning and the availability of powerful computational resources. Since our research objective synthesizes techniques and concepts from numerous fields, the methods are grounded in data science and digital experiments.

Prior research, like Backstrom, Dwork, and Kleinberg \cite{backstrom_wherefore_2007} tackled social network deanonymization, but (like Sweeney) used a dataset that was anonymized after-the-fact. The advantage of these approaches is that the method is fully supervised and has a clear objective. Unfortunately, such methods are not practical against real datasets that achieve their anonymity by cryptographically deriving new addresses. But their methods fit their goals: they propose two graph-based techniques for deanonymization, an active one that requires the prior construction of a subgraph of attacking nodes (prior collusion) to reveal information about neighbours (by walking the graph or making strategic cuts), and a passive method that relies on graph isomorphisms (i.e., `partitioning' rather than `clustering'). Narayanan and Shmatikov’s more recent work on social network deanonymization \cite{narayanan_-anonymizing_2009} draws inspiration from Backstrom et al., but used graph isomorphisms with \emph{auxiliary} graphs to link anonymous identities with public identities, a more information rich context and a method that requires a small number of ‘seed’ nodes present in both graphs. 

Narayanan and Shmatikov’s method for social network deanonymization is effective to the extent their walk-based propagation method can reach unknown nodes. However, since most anonymous networks (like Bitcoin) have many singleton nodes (assuming address rotation best security practices), propagation from known seeds to unknown nodes will not be effective. 

Similarly, more recent work has successfully used an open source information (OSI) approach to identify additional auxiliary graphs for deanonymizing Bitcoin users \cite{blackburn_cooperation_2022}. This research focuses on the ability to walk the transaction graph and uses a ‘follow the money’ approach to track down auxiliary information sources. However, in practical terms, this approach relies on poor security behaviour (high \emph{k}-anonymity) and investigations often lead to an anonymizing coin mixer or a regulated exchange that can only be further investigated by legal authorities.

Recently, there have been significant attempts to deanonymize Bitcoin transactions. Most research uses supervised machine learning techniques to classify addresses, but with the exception of Akcora et al.\ \cite{akcora_bitcoinheist_2019} and Paquet-Clouston et al.\ \cite{paquet-clouston_ransomware_2018}, researchers do not have access to quality labelled datasets. Instead, researchers use synthetic and fake data: Ashfaq et al.\ \cite{ashfaq_machine_2022} use a synthetic dataset; Rabieinejad et al.\ \cite{rabieinejad_generative_2023} generate fake labels; Dahiya et al.\ \cite{dahiya_neural_2023} use an unverified Kaggle dataset; Pham and Lee \cite{pham_anomaly_2017} use unverified labels for ``30 thieves'' of unknown provenance; Sankar Roy et al.\ \cite{sankar_roy_exploiting_2022} use the same Pham and Lee dataset. Or researchers use very simple heuristics (such as node degree patterns, see Weber et al.\ \cite{weber_anti-money_2019} and Lorenz et al.\ \cite{lorenz_machine_2021} who use the Weber et al.\ dataset) or slightly more complex heuristics (like motifs, see Wu et al.\ \cite{wu_detecting_2022}) and assume that such patterns are evidence of complex criminal behaviour, like money laundering. This approach is a methodological dead end; putting aside the assumptions associated with fake, arbitrary, or false datasets, due to the imbalance of classes (very few labels of illegal or abnormal transactions), researchers must use resampling techniques like SMOTE to improve learning performance \cite{wu_detecting_2022}. These are not good data science practices; effectively researchers are using low quality data and stretching it even thinner to boost learning performance (often using overly simple metrics like recall and accuracy). The results have no practical consequence. Indeed, surveying the field with all of these constraints and heuristics in mind, we can conclude that practical Bitcoin deanonymization remains elusive.

The most effective research on deanonymizing Bitcoin transactions uses real labelled datasets and intuitive machine learning algorithms (Akcora et al.\ \cite{akcora_bitcoinheist_2019} and Paquet-Clouston et al.\ \cite{paquet-clouston_ransomware_2018}) or graph deep learning (Zhou et al.\ \cite{zhou_behavior-aware_2022}), but these studies still have significant limitations. Paquet-Clouston et al.\ \cite{paquet-clouston_ransomware_2018} attempt to trace ransomware payments by constructing graphs from 7222 labelled seeds originating from a convenience sample of known ransomware addresses. They construct an address graph and a cluster graph; the latter uses the logic of Sybils to combine nodes that have shared inputs (multiple-input clustering heuristics). However, their analysis is purely descriptive and does not attempt to make predictions. Akcora et al.\ \cite{akcora_bitcoinheist_2019} also investigate ransomware transactions, use convenience sample of three datasets (which includes the Paquet-Clouston et al.\ dataset), and construct a transaction graph and an address graph. They use multiple heuristics, develop graph features, use density-based clustering (DBSCAN) and \emph{k}-means algorithms to cluster addresses with a similarity search. 

Finally, Zhou et al.\ \cite{zhou_behavior-aware_2022} attempt Ethereum address clustering, or what we would call `Sybil identification.' They construct a trimmed graph of Ethereum transactions, develop node features, and train a graph neural network (GNN) to search subgraph node embeddings using a hierarchical (weighted) graph attention autoencoder. As is typical, they use several pooling layers and non-linear activation layers (LeakyReLu). They use cross-entropy loss for the classifier. They improve the performance of the default GNN by using contrastive subgraph learning. This self-supervised methodology, while in a different context, is similar to our approach detailed below. The significant difference is that they classified nodes directly using the GNN, whereas we use develop the embeddings first and then classify using a embeddings search, an encoder-decoder model. The second step is necessary to detect Sybils because Sybils are by definition unknown and with limited connectivity to other voter nodes.

In the context of decentralized governance, however, Sybils are identified using our more relaxed definition that circumvents \emph{k}-anonymity. In this context, our method for identifying Sybils is a weak form of deanonymization but one that reveals little to no personally identifiable information and unlike most previous methods does not require auxiliary or labelled graphs (although it can benefit from such information). Supervised techniques for deanonymization will not effectively identify Sybils because, in a social graph characterized by strong anonymity, Sybils may create a \emph{single}, disconnected edge and thus lack sufficient connectivity to be identified by walk-based deanonymization approaches. Naive representations of Sybils contain only 1 bit of structural information. Indeed, we assume that a majority of Sybils are singletons. The existence of a labelled subgraph of known voters, linked to Ethereum Name Service (ENS) accounts, provides little help in identifying isolated Sybils, for the same reason (but can be used to help validate clusters). 

To overcome this limitation, we define a Sybil as a cluster of `similar' nodes, and singletons are noise. Analytically, this makes sense; if nodes have similar higher dimensional representations and hidden features we can conclude that they are either representative of the same voting person or at least representative of voters acting in concert with one another (perhaps conducting a governance voting attack).

\subsection{Experimental Sybil Identification Methodology}

\begin{figure}[h]
    \centering
    \includegraphics[width=0.6\textwidth]{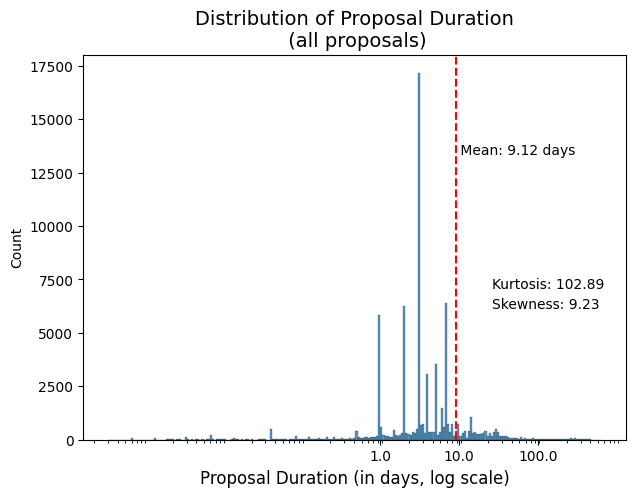}
    \caption{Histogram of unfiltered governance proposal durations.}
    \label{fig:duration_distribution}
\end{figure}

Our dataset covers governance voting for roughly 35,000 DAOs between 2020-07-18 and 2023-02-27 (959 days or 2.6 years), which covers approximately 7.2 million voting transactions. However, naive counts of DAOs, voters, and their governance proposals are ineffective due to noise. For instance, upon manual inspection the dataset reveals many registered DAOs that failed to launch or were clearly set up as tests, and many governance proposals have durations far too short to be practical (see Figure \ref{fig:duration_distribution}).     

We found evidence of extensive wallet address re-use, where a single voter makes multiple votes from the same pseudonymous wallet address (in our dataset, approximately 71\% are reused wallets). In a traditional financial setting, this is considered poor security behaviour (resulting in low \emph{k}-anonymity). However, in a polycentric governance context, this is not necessarily evidence of lax security practices but rather the demands of a different social context.

We performed typical Extract, Transform, and Load (ETL) practices on our dataset. After querying the snapshot.org API, we stored the dataset in a relational database and used a SQL command to order voting transactions by date. After sanitizing and checking the data, we constructed the voting graph. On the voting graph, we set node and edge attributes, including the amount of `voting power' used for each vote on a proposal, UNIX timestamps, and persistent IDs linked to Ethereum Name Service IDs, when available. The resulting voting graph constructed from the snapshot.org dataset is available for inspection and statistical analysis (see \hyperlink{https://github.com/quinndupont/SybilGovernance}{Github}).

\subsubsection{`Follow the Money' Sybil Chain Subgraph Search}

Our dataset captures the ``off-chain'' voting activities of thousands of DAOs on the governance platform snapshot.org (setting these DAOs apart from more autonomous DAOs, like MakerDAO, that use ``on-chain'' voting to directly modify organizational parameters). Tracing a token vote across its origin chain would clearly be informative and useful for our deep learning method, but this feature proves challenging to engineer. Recall that off-chain voting exists in large part to address high gas fees associated with main chain transactions (Ethereum or otherwise); thus, off-chain voting leaves no transaction chains to follow. Instead, the process of identifying a ``Sybil chain'' would require pinpointing the specific token(s) `used' at the time of the vote (used in the sense that they are counted by specific voting `strategies'), induce an \emph{n-}deep path on the origin chain, and conduct a breadth-first search to connect two or more previously unknown voters. However, despite its potential to enhance our self-supervised approach, the problems associated with walking origin graphs are significant, such that implementation would entail addressing a distinct set of issues and a different research agenda.

Thus, naively tracing token transactions is thwarted by robust privacy protections inherent in the various origin chains. To adapt to these constraints and for experimental purposes, we insert a randomly generated fake Sybil in the dataset. This controlled element is then identified as part of our testing protocol. The inclusion of this artificial Sybil serves as a test case, assessing the efficacy of our methods under controlled conditions and providing insights into the potential and limitations of our approach in dealing with real-world scenarios.

\begin{figure}[h]
    \centering
    \includegraphics[width=0.6\textwidth]{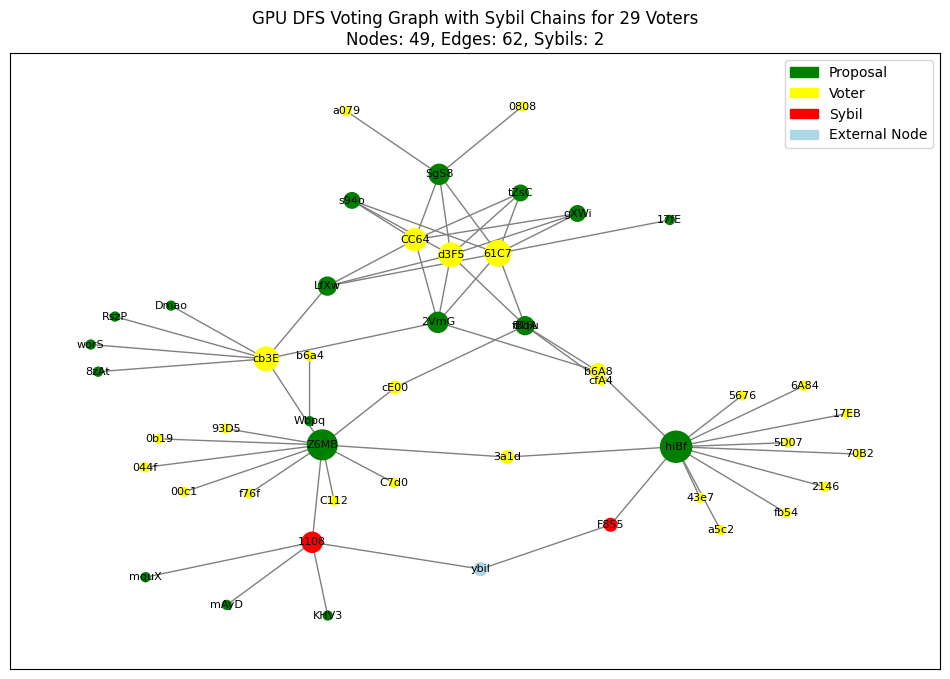}
    \caption{Small voting graph with fake Sybil chain identified using BFS.}
    \label{fig:Sybilchain}
\end{figure}

However, variants of this `follow the money' approach \cite{blackburn_cooperation_2022} have been successfully developed, refined, and even commercialized. For instance, although no longer maintained, \hyperlink{https://github.com/citp/BlockSci}{BlockSci} was the benchmark for Bitcoin taint analysis and forensics; today, many commercial platforms offer forensics capabilities (\hyperlink{https://www.chainalysis.com/}{Chainalysis} is currently a market leader).

Even in the `simplest' case of Bitcoin, privacy preserving blockchains present a number of obstacles to forensic and especially taint analysis. Rather straightforwardly, since Bitcoin uses an unspent transaction output (UTXO) accounting model, it is possible to determine when wallet addresses that require the exact same knowledge to spend can be considered equivalent (BlockSci's ``equivalent addresses''), which is a type of direct Sybil identification. However, even when assuming users are \emph{not} following privacy-maximizing address rotation practices (about half of Bitcoin transactions), tracing financial flows still requires linking inputs and outputs, addressing ``change'' wallets, and surmounting coinJoin transactions and coin mixing services.

By combining multiple heuristics with supervised classification tasks, \cite{kappos_how_2022} were able to improve link prediction across Bitcoin transactions, besting BlockSci's static and dynamic heuristics. Indeed, Meiklejohn's corpus of research on the co-spend (or multiple input) clustering heuristic \cite{meiklejohn_fistful_2013} and more complex ``peel chains'' \cite{kappos_how_2022} exemplify the practical challenges of linking Bitcoin transactions. In another recent work, Möser and Narayanan \cite{moser_resurrecting_2022} improve on earlier de-anonymizing techniques by refining existing heuristics (including multiple input and change transactions) and using a Random Forest classifier to cluster similar transactions. Like ours, their method uses self-supervised learning (bootstrapped from a collection of heuristics), but uses a more conventional --- lower dimensional --- clustering method. In addition to heuristics, they engineer a number of fingerprinting features. They report impressive ROC scores but the method is limited by the numerous dataset exclusions required to prevent cluster collapse (e.g., singletons cannot be identified). Notably, this form of ``address clustering'' is simply another definition of Sybil identification. And finally, while none of these approaches attempt to dissect coinJoin and mixing techniques, the \hyperlink{https://code.samourai.io/oxt/boltzmann}{`Boltzmann' coefficient} measures the difficulty of predicting links by computing the entropy of Bitcoin transactions. A similar research challenge remains for Ethereum and other chains.

\subsubsection{Graph Time Signal Analysis}
For our dataset of 7.2m votes, using the PyGSP Python library, we constructed several wavelet graph signal filters to better understand the frequency components of the voter time series. Wavelet transforms offer a powerful mechanism for decomposing time series data into a spectrum of frequencies across different scales, enabling the extraction of nuanced, multi-scale features that encapsulate both the frequency and temporal information inherent to the signal

\begin{figure}[h]
    \begin{minipage}{0.48\textwidth}
        \centering
        \includegraphics[width=1\textwidth]{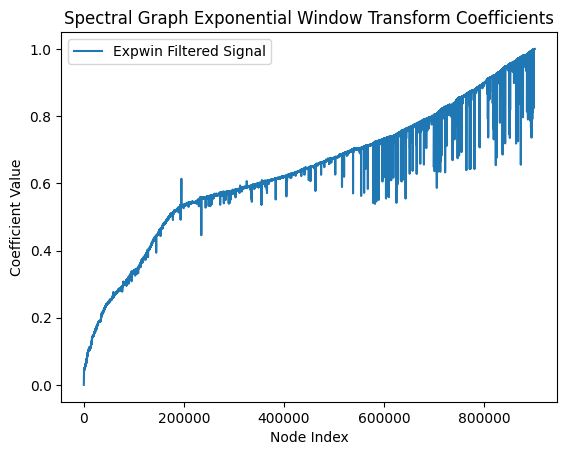}
        \caption{Expwin Filter Histogram.}
        \label{fig:expwin_filter}
    \end{minipage}
    \hfill 
    \begin{minipage}{0.48\textwidth}
        \centering
        \includegraphics[width=1\textwidth]{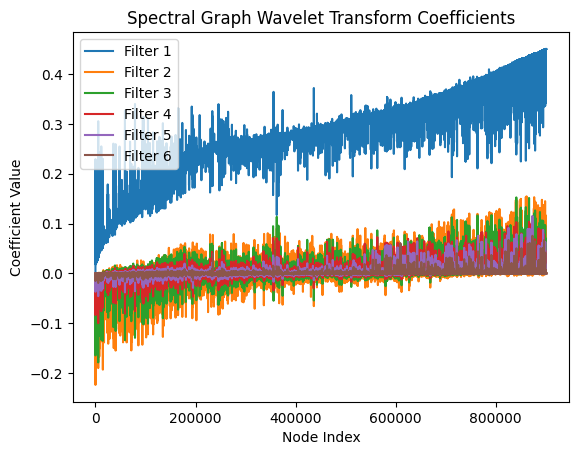}
        \caption{Spectral Graph Wavelet Transform Histogram.}
        \label{fig:sgw_filter}
    \end{minipage}
\end{figure}

First, the Expwin filter, characterized by an exponential decay, is a low-pass filter with a frequency response expressed as:
\[
g_{\text{exp}}(\lambda) = e^{-\tau \lambda}
\]
where $\lambda$ are the eigenvalues of $L$, and $\tau$ controls the decay rate. This filter emphasizes the low-frequency components of the signal, ideal for denoising and highlighting global trends across the graph (see Figure \ref{fig:expwin_filter}).

Second, Spectral Graph Wavelet Transform (SGWT) is used to analyze graph signals by computing the Laplacian operator $L$, which encodes the connectivity structure of the graph, and its eigenvalues and eigenvectors, analogous to the classical Fourier transform. When a signal $x$ on a graph undergoes SGWT using a filter $g_i$, it is decomposed into components corresponding to different frequency bands captured by the wavelet filters (see Figure \ref{fig:sgw_filter}). Mathematically, the transform is given by:
\[
c_i = g_i(L) x = U g_i(\Lambda) U^T x
\]
where $c_i$ are the coefficients after applying the $i$-th filter, $U$ is the matrix of eigenvectors of $L$, $\Lambda$ is the diagonal matrix of eigenvalues of $L$, and $g_i(\Lambda)$ applies the filter in the spectral domain.

\begin{figure}[h]
    \centering
    \includegraphics[width=0.6\textwidth]{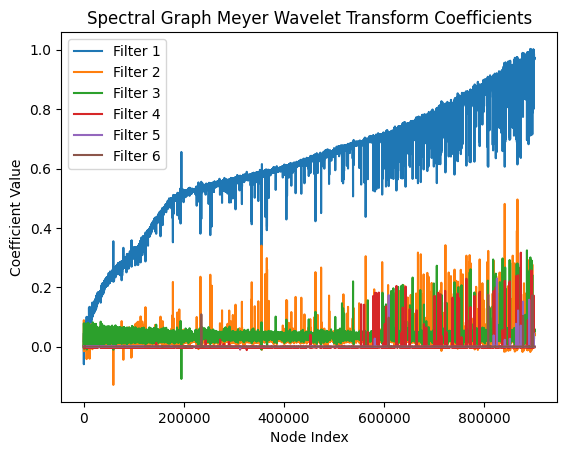}
    \caption{Spectral Graph Wavelet Transform Histogram.}
    \label{fig:meyer_filter}
\end{figure}

And finally, the Meyer Wavelet filter bank provides multi-resolution analysis and is a tight frame, suitable for decomposing signals into different scales. Its general form in the spectral domain is:
\[
g_{\text{Meyer}}(\lambda) = \nu\left(\frac{\lambda}{\lambda_{\text{max}}}\right)
\]
where $\nu(\cdot)$ transitions smoothly between 0 and 1, typically using sine and cosine functions. It captures both global and local features, valuable for feature extraction and signal compression (see Figure \ref{fig:meyer_filter}).

The Expwin filter is best for capturing overall smoothness, while the Meyer wavelet strikes a balance between smoothness and localization. The SGWT with filters helps describe the signal's structure on the graph. High coefficients in the SGWT indicate regions where the signal strongly resonates with the filter, pointing to significant features at different frequency bands, which can be exploited by deep learning. 

In the pursuit of elucidating the complex temporal dynamics of DAOs, we engineer and embed spectral time series data, use wavelet transforms with unsupervised clustering techniques to reduce the high dimensional signal. This multi-resolution analysis is particularly useful for highlighting non-linear patterns and dynamics obscured in raw time series data, making wavelets a suitable preprocessing step for uncovering spectral time embeddings.

Following the wavelet transformation, clustering algorithms are used to group vectors based on similarity of their spectral and temporal features, revealing distinct patterns or behaviors within the dataset. Clustering is important for understanding the inherent structure and relationships within the data, providing a basis for further analysis and visualization. Finally, reducing the high-dimensional clusters into a lower-dimensional space preserves the essential topological structure of the data. This dimensionality reduction enables a visual exploration of the clustered spectral time embeddings and underscores the relationships and distinctions between identified patterns. The amalgamation of wavelet-based feature extraction, clustering, and visualization presents an experimental framework for analyzing time series data, helping unearth and visualize complex spectral time dynamics that would otherwise remain concealed.

\subsubsection{Wallet Fingerprint Signal Analysis}
\begin{figure}[h]
    \begin{minipage}{0.48\textwidth}
        \centering
        \includegraphics[width=\linewidth]{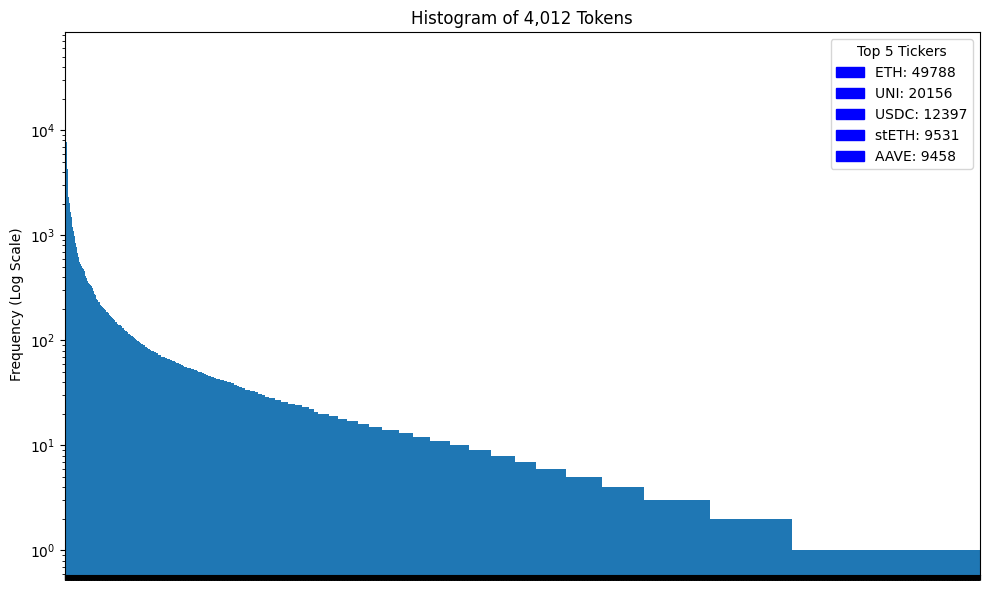}
        \caption{Histogram of Unique Wallet Tokens.}
        \label{fig:uniquetokens}
    \end{minipage}
    \hfill 
    \begin{minipage}{0.48\textwidth}
        \centering
        \includegraphics[width=\linewidth]{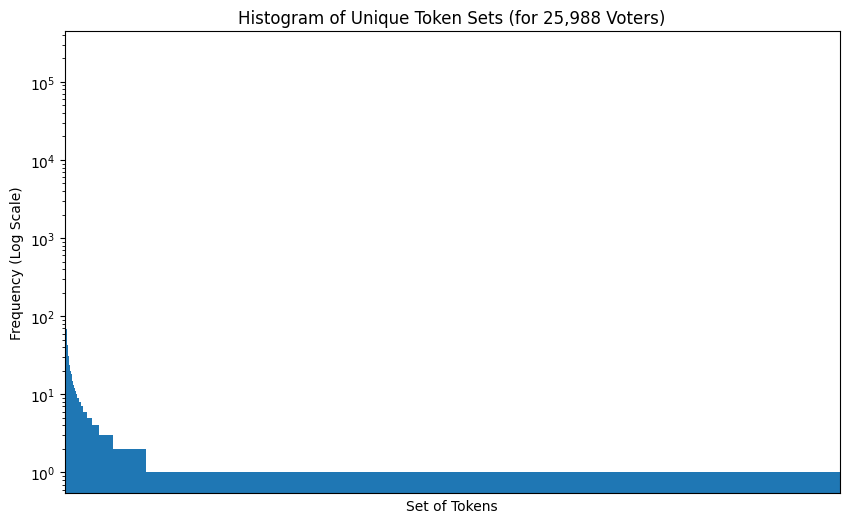}
        \caption{Histogram of Unique Wallet Token Sets.}
        \label{fig:wallettokensets}
    \end{minipage}
\end{figure}

\begin{figure}[h]
    \centering
    \includegraphics[width=0.8\textwidth]{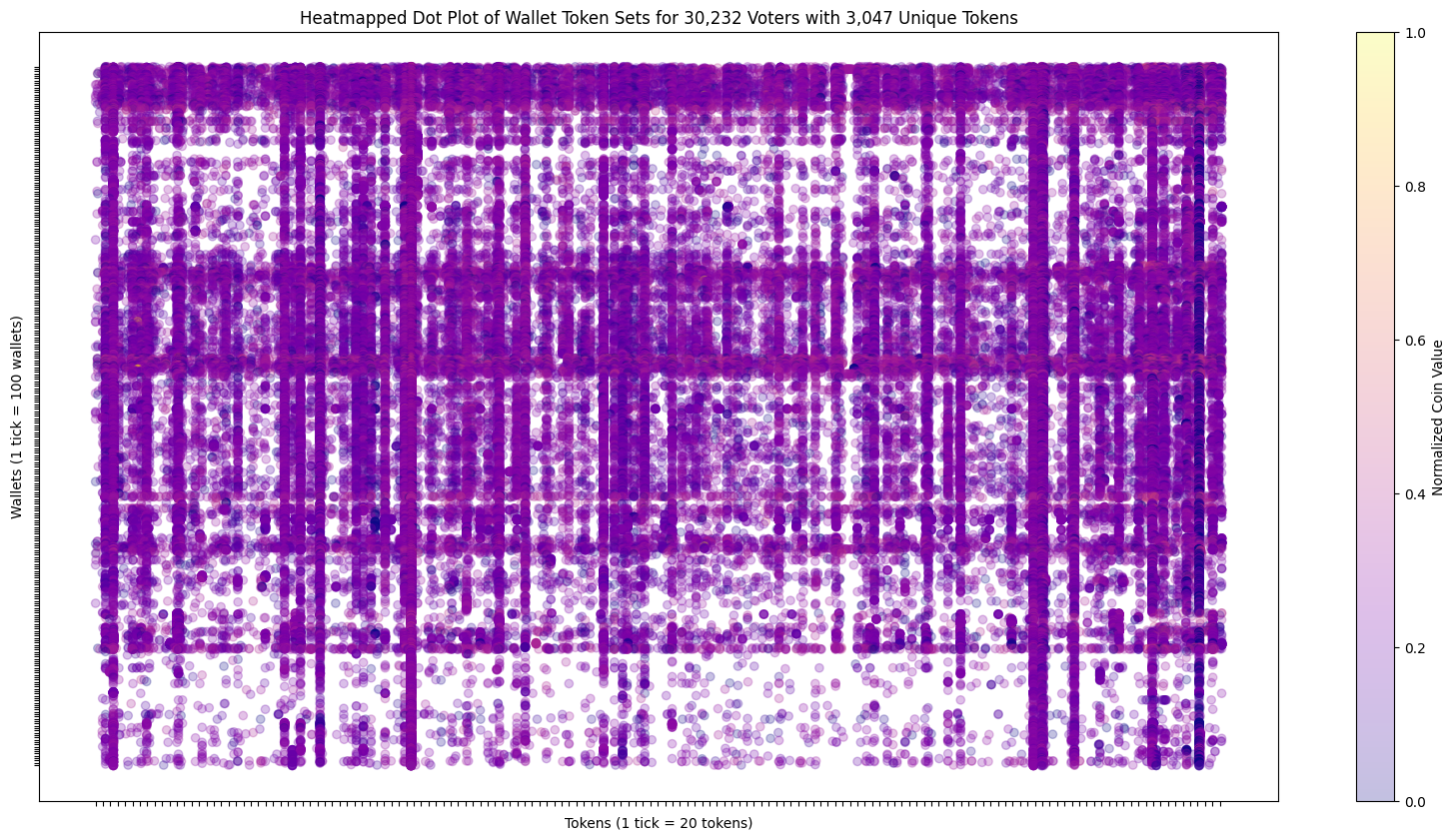}
    \caption{Dot Plot of Voter Wallet Token Sets with Colormapped Values.}
    \label{fig:dotplot}
\end{figure}

While we cannot directly validate Sybils by tracing origin chain transactions, we can use historical wallet balances to ``fingerprint'' voters. Using the snapshot.org API, we constructed a graph of voters for seven DAOs of varying sizes and augmented each voter with unique wallet token balances at the time of the vote. We used the Covalent Unified API to collect token tickers and their associated historical balances and attached these `coins' to each voter. From Figures \ref{fig:uniquetokens} and \ref{fig:wallettokensets}, we see many unique tokens in voter wallets and observe that voters share wallet token sets, with a long tail of shared, distinct voter sets. If we then dot plot the token sets and colormap their normalized values, we see clear evidence of extractable information for each wallet defined by the uniqueness of wallet token sets (Figure \ref{fig:dotplot}). 

\begin{figure}[h]
    \centering
    \includegraphics[width=0.6\textwidth]{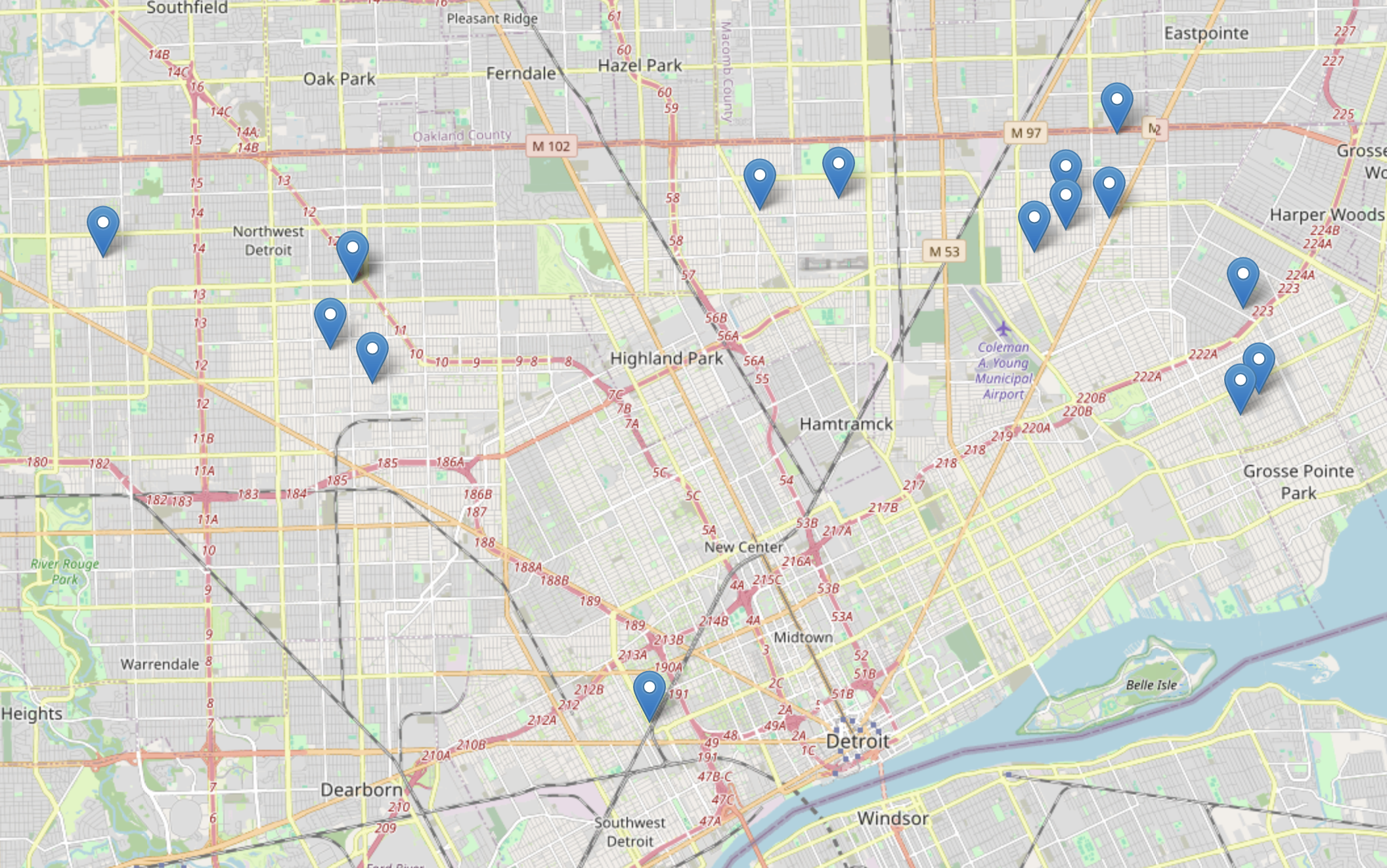}
    \caption{Map of REALTOKENS in Wallet of Anonymous Voter.}
    \label{fig:realtokensmap}
\end{figure}

Indeed, some token sets are highly individual, like the `anonymous' voter who owns a set of \hyperlink{https://realt.co/}{REALTOKENS} (a form of tokenized, fractional property ownership) with specific addresses in Detroit, Michigan (see Figure \ref{fig:realtokensmap}:
\begin{enumerate}
\item {18481 WESTPHALIA ST DETROIT MI}
\item {9717 EVERTS ST DETROIT MI}
\item {13895 SARATOGA ST DETROIT MI}
\item {18466 FIELDING ST DETROIT MI}
\item {15095 HARTWELL ST DETROIT MI}
\item {...}
\end{enumerate}

\noindent More practically, we also analyse the wallet fingerprints using Uniform Manifold Approximation and Projection (UMAP), a manifold learning technique for dimensionality reduction. UMAP seeks to reduce \( D \) to a lower dimension \( d \), resulting in \( \mathbf{E}_{\text{reduced}} \in \mathbb{R}^{N \times d} \), while preserving the local and global structure of the data. While the UMAP algorithm is quite simple (it calculates nearest-neighbours to produce sets of weighted graphs, calculates probabilities for edges on a unified graph, and then uses a force algorithm to lay the graph out), the mathematical underpinnings are fascinating and much more complex \cite{mcinnes_umap_2020}. In a sentence, UMAP finds fuzzy simplicial sets (mathematical abstractions from category theory that involve objects and mappings), which define the metric between nodes (viz.~Riemannian in that space is made of individually defined metrics, unlike Euclidean space); to display a high dimensional representation in low (2 or 3) dimensions, UMAP minimizes the cross-entropy between the high-dimensional fuzzy simplicial set and the \emph{a priori} low dimensional manifold.

\begin{figure}[h]
    \begin{minipage}{0.48\textwidth}
        \centering
        \includegraphics[width=1\textwidth]{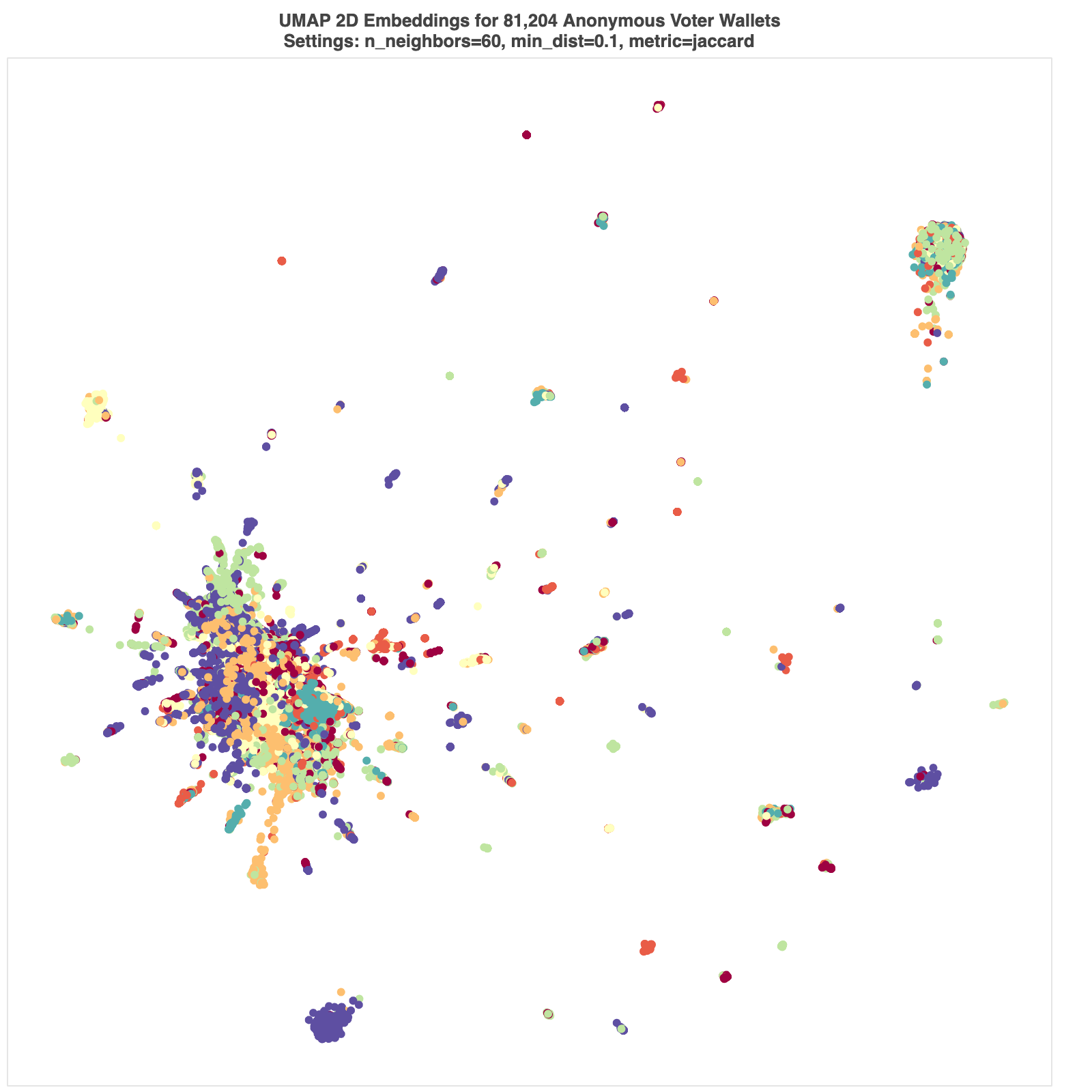}
        \caption{UMAP Projection of Wallet Token Fingerprints.}
        \label{fig:2dembeddings}
    \end{minipage}
    \hfill 
    \begin{minipage}{0.48\textwidth}
        \centering
        \includegraphics[width=1\textwidth]{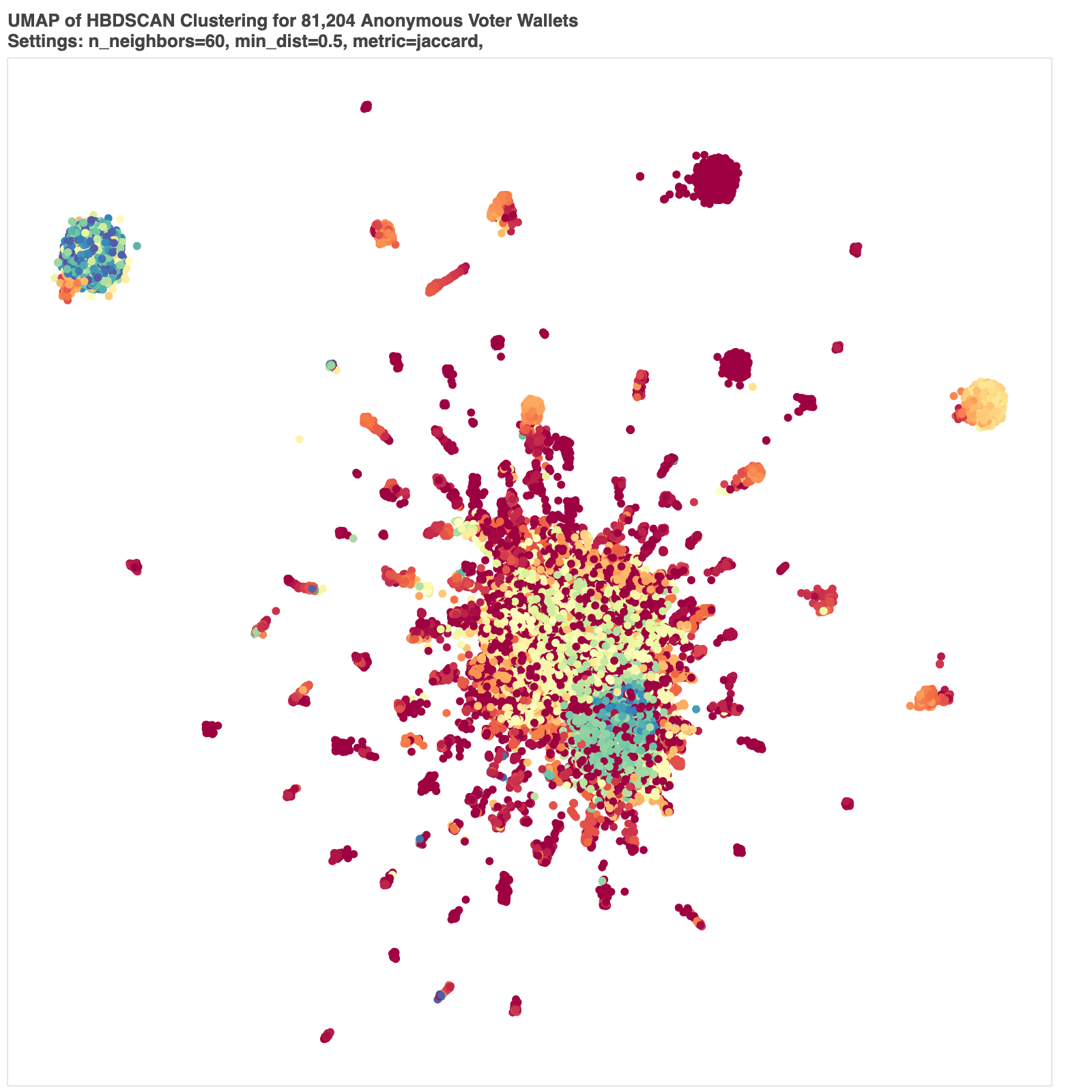}
        \caption{UMAP Projection of HDBSCAN Clustered Wallet Token Fingerprints.}
        \label{fig:2dclusteredHDBSCAN}
    \end{minipage}
\end{figure}

To prepare our fingerprint dataset, we encode each voter's unique ticker set as a sparse matrix. We then fit the wallet's vector embeddings using a Jaccard metric with UMAP and construct an interactive 2D or 3D plot, which reveals features of the voter fingerprints (the interactive plot is available on \hyperlink{https://github.com/quinndupont/SybilGovernance}{Github}). Through experimentation we found that the Jaccard, cosine, and Manhattan metrics all work reasonably well with a default 15 neighbours, but Jaccard offered the best balance of local and global topology (see Figure \ref{fig:2dembeddings}).

\begin{figure}[h]
    \centering
    \includegraphics[width=0.6\textwidth]{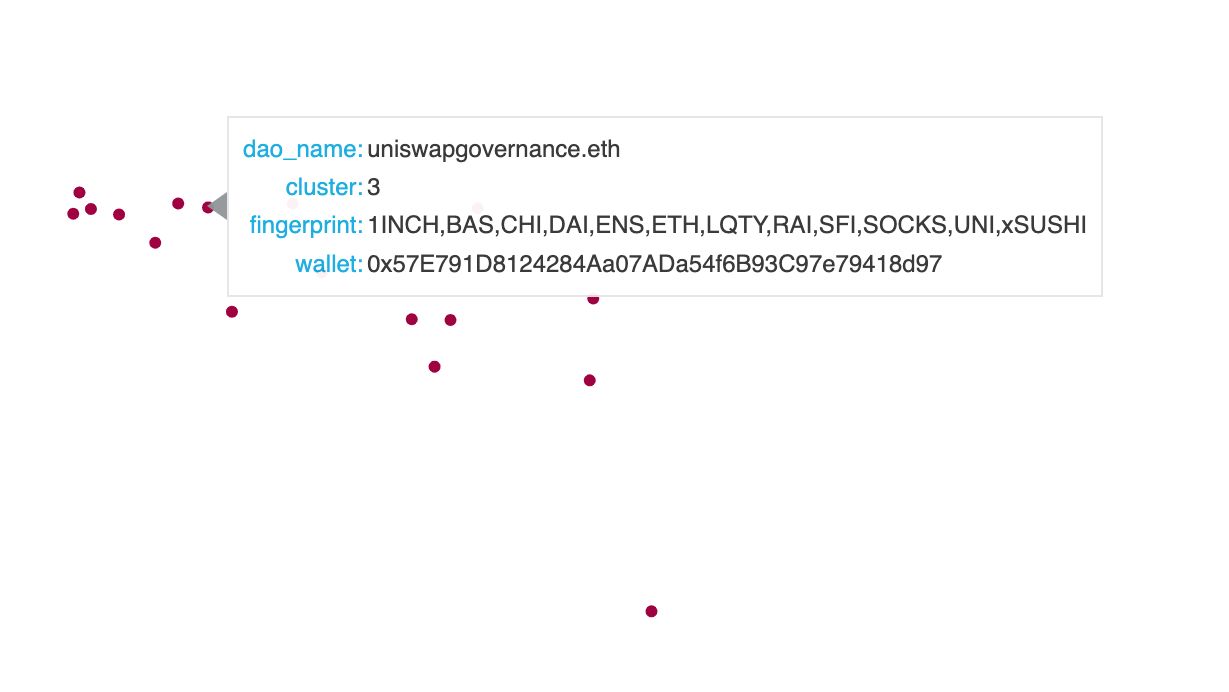}
    \caption{Suspected Sybils Determined by Wallet Fingerprints (nodes share a highly distinct fingerprint).}
    \label{fig:Sybilfingerprints}
\end{figure}

We then cluster the embeddings using Hierarchical Density-Based Spatial Clustering of Applications with Noise (HDBSCAN) and refit the embeddings with UMAP to visualize the results (see Figure \ref{fig:2dclusteredHDBSCAN}). Inspecting the plot is revealing; for instance, when tight, uniform clusters are far from others, we suspect the presence of Sybils (clusters reveal sets of users with a highly unique wallet fingerprint; see Figure \ref{fig:Sybilfingerprints}.

Finally, we train a simple deep neural network to learn each voter's wallet token fingerprint and predict the associated DAO. While there is little practical value in this prediction, it shows that wallet sets contain important topological information. Indeed, our clustering results have an accuracy of 0.6, which is significantly better than random guessing (0.14) for seven classes, by 45.7\%, although there is clearly much room for improvement.  

Specifically, given a dataset \( \mathcal{D} = \{(\mathbf{x}_i, y_i)\}_{i=1}^{N} \) where \( \mathbf{x}_i \in \mathbb{R}^{d} \) represents the input features of dimension \( d \) and \( y_i \) represents the labels. The dataset is divided into training, validation, and testing sets. Let \( f(\mathbf{x}; \theta) \) be the neural network function parameterized by \( \theta \).

The neural network, \( f(\mathbf{x}; \theta) \), is defined as follows:
\begin{enumerate}
    \item The input layer takes \( d \)-dimensional input vector \( \mathbf{x} \).
    \item The first linear transformation is applied: \( \mathbf{z}_1 = \mathbf{W}_1\mathbf{x} + \mathbf{b}_1 \), where \( \mathbf{W}_1 \in \mathbb{R}^{512 \times d} \) and \( \mathbf{b}_1 \in \mathbb{R}^{512} \).
    \item A ReLU activation function is applied: \( \mathbf{a}_1 = \text{ReLU}(\mathbf{z}_1) \).
    \item Batch normalization is applied to \( \mathbf{a}_1 \) before dropout: \( \mathbf{a}_1' = \text{BatchNorm}(\mathbf{a}_1) \).
    \item Dropout is applied to \( \mathbf{a}_1' \) with a dropout rate of 0.5.
    \item This process is repeated for additional layers with dimensions \( 512 \to 256 \to 128 \to 64 \), including linear transformations, ReLU activations, batch normalization, and dropout at each step.
    Specifically, for layer \( l \) where \( 2 \leq l \leq 5 \):
   
   \[
   \mathbf{z}_l = \mathbf{W}_l\mathbf{a}_{l-1}' + \mathbf{b}_l, \quad \mathbf{a}_l' = \text{Dropout}(\text{BatchNorm}(\text{ReLU}(\mathbf{z}_l)), 0.5)
   \]
   
    with \( \mathbf{W}_l \) and \( \mathbf{b}_l \) representing the weights and biases of layer \( l \), respectively.
    \item The final layer's output \( \mathbf{z}_5 \) is passed through a linear transformation without a subsequent activation or dropout for the classification: \( \hat{\mathbf{y}} = \mathbf{W}_5\mathbf{a}_4' + \mathbf{b}_5 \), where \( \mathbf{W}_5 \in \mathbb{R}^{\text{num\_classes} \times 64} \) and \( \mathbf{b}_5 \in \mathbb{R}^{\text{num\_classes}} \).
\end{enumerate}

He initialization is applied to all linear layers, ensuring variance scaling is maintained in the network's activations. The network is trained using the cross-entropy loss function: \[
\mathcal{L}(\theta) = -\frac{1}{N} \sum_{i=1}^{N} \log \frac{e^{\hat{y}_{i,y_i}}}{\sum_{j=1}^{\text{num\_classes}} e^{\hat{y}_{i,j}}}
\] where \( \hat{y}_{i,j} \) is the \( j \)-th element of the network's output \( \hat{\mathbf{y}}_i \) for input \( \mathbf{x}_i \), and \( y_i \) is the true label of \( \mathbf{x}_i \).

Optimization is performed using the Adam optimizer with a learning rate of \( 0.001 \) and weight decay of \( 1e-5 \) for regularization, and the model is evaluated using a split of training, validation, and testing data.

\begin{figure}[h]
    \begin{minipage}{0.48\textwidth}
        \centering
        \includegraphics[width=1\textwidth]{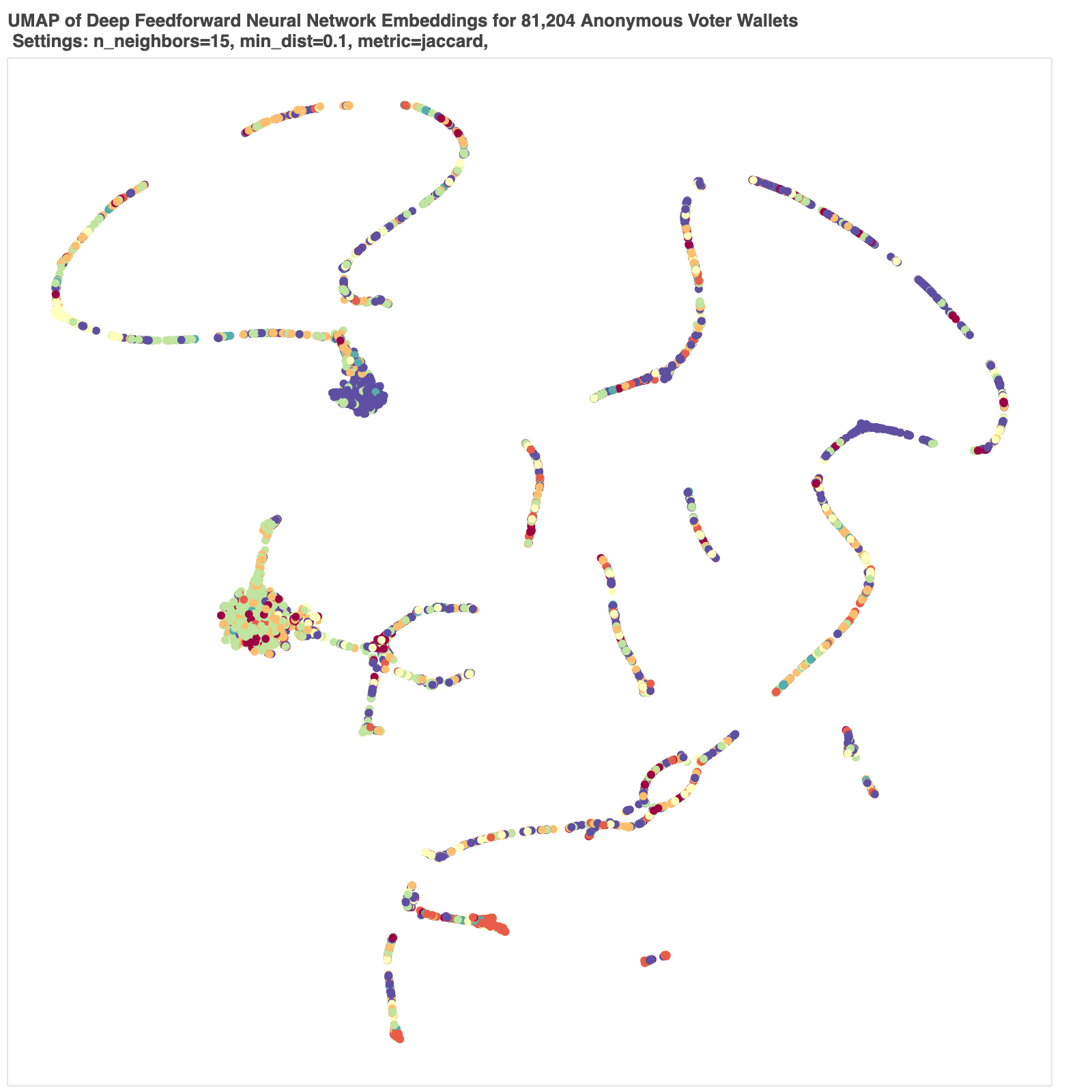}
        \caption{UMAP Projection of Deep Feedforward Neural Network Embeddings of Wallet Token Fingerprints.}
        \label{fig:nn_embeddings_UMAP}
    \end{minipage}
    \hfill 
    \begin{minipage}{0.48\textwidth}
        \centering
        \includegraphics[width=1\textwidth]{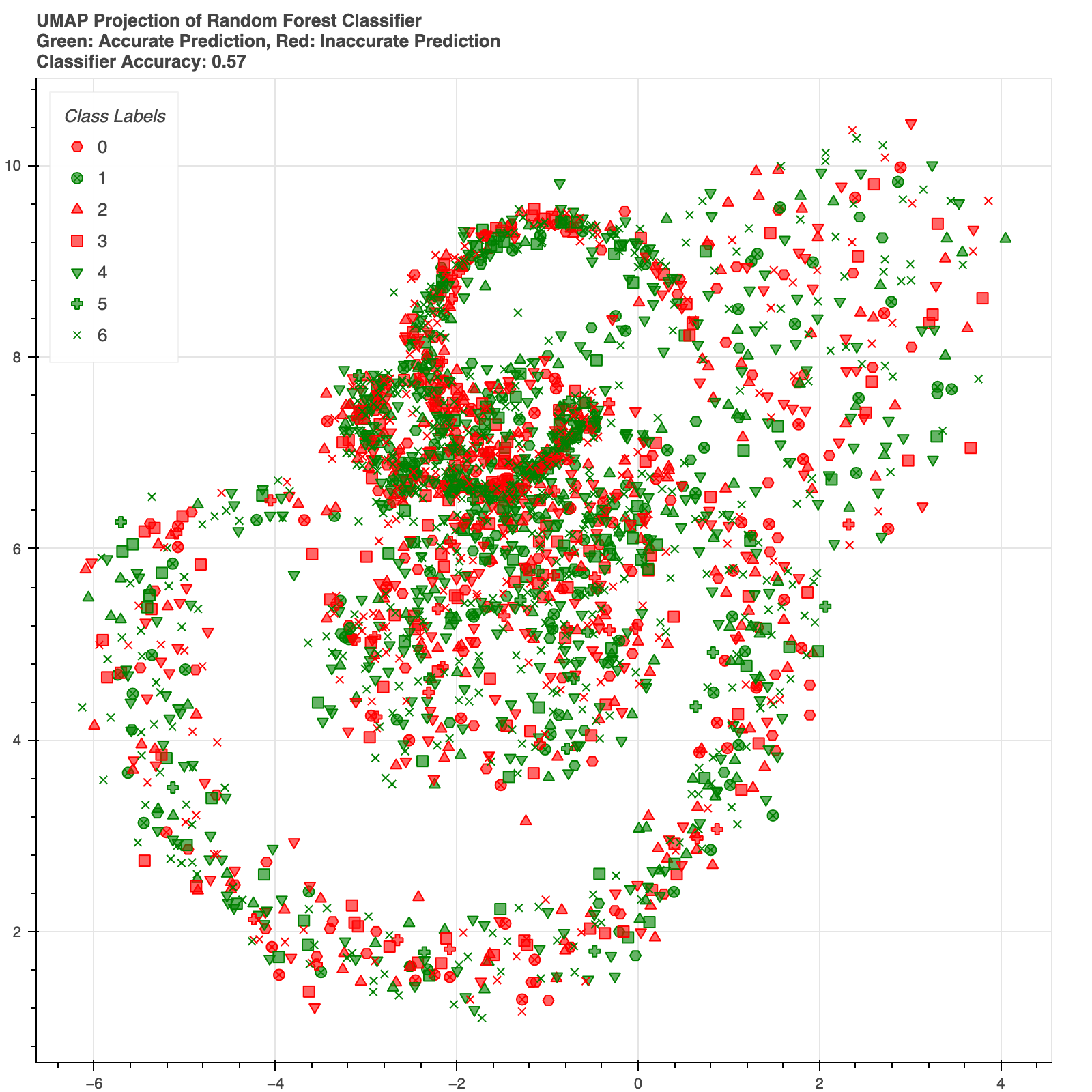}
        \caption{UMAP Projection of Random Forest Clustering of Wallet Token Fingerprints (Colored by Prediction Accuracy).}
        \label{fig:rf_cluster_UMAP}
    \end{minipage}
\end{figure}

After training, the dimensionality of embeddings is once again reduced using UMAP for inspection. We see structure, but the results are of limited value (see Figure \ref{fig:nn_embeddings_UMAP}). Instead, we cluster these high dimensional embeddings using a Random Forest classifier (see Figure \ref{fig:rf_cluster_UMAP}).

\begin{figure}[h]
    \centering
    \includegraphics[width=0.6\textwidth]{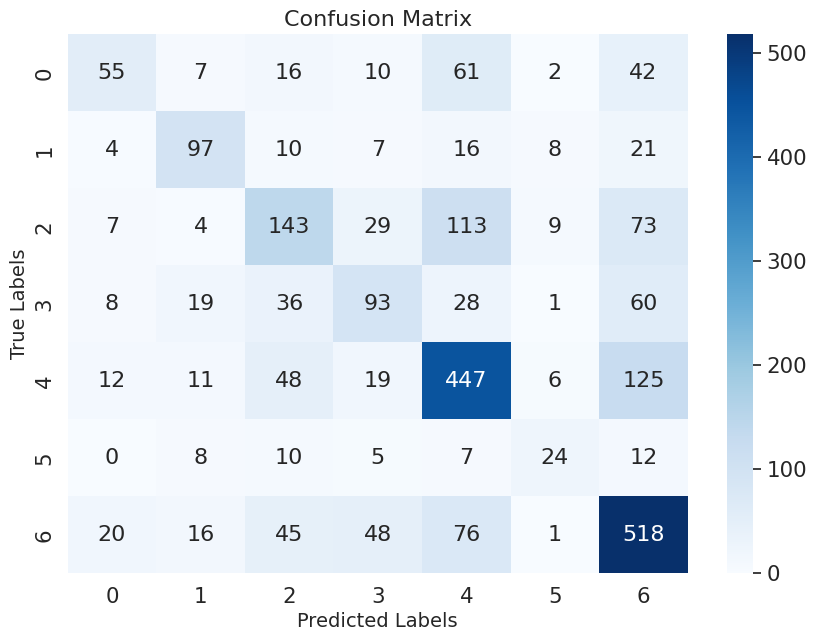}
    \caption{Confusion Matrix of Clustering of Wallet Token Fingerprints.}
    \label{fig:confusionmatrix}
\end{figure}

\noindent The resulting clusters can be evaluated against the true labels with a confusion matrix (Figure \ref{fig:confusionmatrix}). Additionally, we computed cluster evaluation metrics (see Table \ref{table:clustering_evaluation_metrics}).

\begin{table}[h]
\centering
\begin{tabular}{lcccc}
\hline
Class & Precision & Recall & F1-Score & Support \\ 
\hline
0 & 0.52 & 0.28 & 0.37 & 193 \\
1 & 0.60 & 0.60 & 0.60 & 163 \\
2 & 0.46 & 0.38 & 0.42 & 378 \\
3 & 0.44 & 0.38 & 0.41 & 245 \\
4 & 0.60 & 0.67 & 0.63 & 668 \\
5 & 0.47 & 0.36 & 0.41 & 66 \\
6 & 0.61 & 0.72 & 0.66 & 724 \\
\hline
\multicolumn{5}{l}{Accuracy: 0.5650 (2437 samples)} \\
\multicolumn{5}{l}{Macro avg: 0.53 precision, 0.48 recall, 0.50 F1-score} \\
\multicolumn{5}{l}{Weighted avg: 0.55 precision, 0.57 recall, 0.55 F1-score} \\
\hline
\end{tabular}
\caption{Clustering Evaluation Metrics}
\label{table:clustering_evaluation_metrics}
\end{table}

While there remains much room for methodological improvement, our analysis reveals the predictive power of wallet token fingerprinting.

\subsubsection{Graph Convolutional Neural Network (GCNN)}

With a large voting graph, we engineer learnable features and train a Graph Convolutional Neural Network (GCNN) to create vector embeddings for each node and its edge connections. Since there are any number of possible custom GCNNs tailored for self-supervised node classification tasks on graphs, each with a multitude of interchangeable layers for various tasks, we designed the mechanism inductively with iterative testing against the dataset. We proceed experimentally and use any tools and techniques that work well. For instance, above we used a deep feedforward neural network architecture to learn the parameters and then clustered with a Random Forest, which also worked well. In general, as Simon Prince notes \cite{prince_understanding_2024}, most overparameterized deep models work well, despite vast differences in algorithm architecture and mathematical approach. 

For our method, the first few steps are relatively straightforward. First, construct a voting graph where known (labelled) and unknown voters are combined into single nodes with multiple edges to proposal nodes (we use a multi- or hyper-graph to accommodate multiple edges from each node). Second, we engineer features to learn higher dimensional vector embeddings on all nodes (this is the encoder). We use a simple reconstruction loss (Mean Squared Error or MSE) to capture node features for training the GCNN (we do not use a more conventional classification training loss like cross-entropy because we do not know the Sybil classes \emph{a priori}). Third, we use the node embeddings to conduct a \emph{k}-means vector clustering on unknown nodes (this is the decoder). We then propagate the most common label associated with each cluster to relabel the actual, lower dimensional nodes to form the Sybil cluster (this is also a type of MaxPooling layer for dimensionality reduction). Finally, we take the relabelled similarity graph and combine nodes with their labels in each cluster to form a higher dimensional, but reduced-size graph (upscaling). The resulting graph is an inductive prediction of Sybil nodes.

The deep learning method was developed with the Python libraries NetworkX and RAPIDS AI to construct the graphs, the PyTorch Geometric library for the GCNN, and the Facebook AI Similarity Search (FAISS) for the fast embeddings clustering. Whenever possible, GPU accelerated libraries were used, although speed and memory tradeoffs quickly become a practical data analysis challenge when dealing with large graphs requiring several hundred gigabytes of memory for large matrix multiplications during learning. This is known as the graph expansion problem and can be controlled by neighborhood sampling or graph partitioning (in our case, batch normalization). As an inductive method, decisions to use a particular library, feature, or machine learning algorithm were made pragmatically and tested against the dataset, but as discussed below, the GCNN autoencoder approach offers some distinct conceptual advantages over other approaches.

We develop features and construct a typical train/test/validation split for the forward pass of our GCNN. The first layer of the forward pass is fully connected, capturing node features and incorporating voting edges with their ‘power’ attributes. The second layer integrates edge and node features with a sequential multi-layer perceptron (MLP) using ReLu activation, which is then fed into a convolutional layer with mean feature aggregation. The third layer processes temporal sequences engineered for a Long Short-Term Memory (LSTM) layer. Finally, the fourth layer uses a multi-headed Graph Attention Network (GAT) to better learn and search edge connectivity. Due to the final attention layer, this encoder can be considered a transformer with residual connections. (Note that due to dataset availability limitations, the wallet fingerprint signal was not used in this final algorithm.) 

Each of these layers has tuneable hyperparameters that were set by conducting a grid search. After each forward pass and backpropagation to set training weights, training loss (MSE) was calculated, visualized, and assessed. The resulting tensors were normalized on zero and were checked for sufficient numeric variability.

Summarizing, the four layer GCNN to identify Sybils:
\begin{enumerate}
    \item Fully Connected (FC) Layer with voting power
    \item Multi-Layer Perceptron (MLP) Convolutional Layer with node features
    \item Long Short Term Memory (LSTM) with temporal features
    \item Multi-Headed Graph Attention Network (GAT) with edge index.
\end{enumerate}

\subsubsection{Fast Vector Embeddings Search using FAISS}

Once we have learned higher dimensional voter and voting representations for all (known and unknown) nodes, we can cluster similar vector embeddings (an encoding task), which we initially learned using the GCNN. In deciding on our method, we also considered other similarity metrics like Euclidian distance, cosine similarity, and Jaccard similarity, but computing many-to-many similarity searches on large matrixes is very slow; \(O(n^2 \cdot d)\). Thankfully, the Facebook AI Similarity Search (FAISS) Python library provides extremely fast and easily scaled search and clustering options (including GPU processing) for vector embeddings. FAISS offers several search index options for improving efficiency, including approximate search methods that can scale to billions of elements. For our purposes, the L2 index was quick enough and provides exact results (for much larger datasets, FAISS provides approximate methods that are substantially quicker). We use the included \emph{k}-means clustering method to find Sybil clusters, which uses centroids in the high dimensional embeddings to cluster and return neighbours. We normalize the clusters by dropping any remaining singletons and large clusters (mean + 1 standard deviation). 

\begin{figure}[h]
    \centering
    \includegraphics[width=0.6\textwidth]{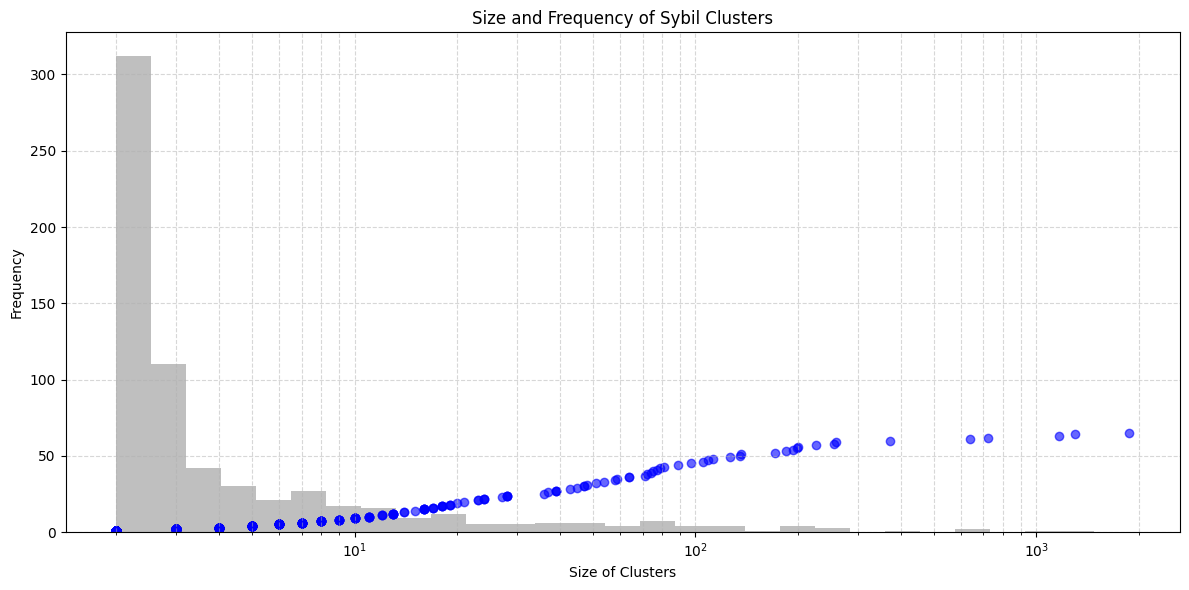}
    \caption{Normalized Sybil Cluster Sizes.}
    \label{fig:clusterdistribution}
\end{figure}

We then inspect each cluster and propagate the most common label across the cluster to construct a ``similarity'' graph (downsampling). Finally, we construct a reduced ``clustered'' graph by combing clustered nodes and reconstructing edges (upsampling).

\begin{table*}[ht]
\centering
\begin{tabular}{lr}
\hline
\textbf{Description} & \textbf{Value} \\
\hline
Total clusters formed (excluding singletons) & 655 \\
Average cluster size & 131.95 nodes \\
Largest cluster contains & 41,149 nodes \\
Smallest cluster contains & 2 nodes \\
Filtered clusters formed (excluding large clusters) & 651 \\
Largest cluster after filtering contains & 1,877 nodes \\
\hline
\end{tabular}
\caption{Summary of Cluster Analysis for 300,000 Votes}
\label{tab:cluster_summary}
\end{table*}

In formal notation, the complete experimental Sybil identification method:

\begin{itemize}
  \item $X$: Node features matrix (degree features).
  \item $E$: Edge features matrix (voting power).
  \item $T$: Temporal features matrix.
  \item $A$: Adjacency matrix of the graph.
  \item $W_{\text{FC}}, b_{\text{FC}}$: Weights and bias of the Fully Connected layer.
  \item $W_{\text{NN}}, b_{\text{NN}}$: Weights and bias of the NNConv layer.
  \item $W_{\text{LSTM}}, h$: Weights and hidden state of the LSTM layer.
  \item $W_{\text{GAT}}$: Weights of the GAT layer.
  \item $D$: Embedding dimension for FAISS L2 index.
\end{itemize}

The GCNN model can be expressed as:
\begin{align*}
\text{GCN}(X, A, E, T, P) &= \text{GAT} \\
&\quad \Big(\text{LSTM}\Big(\text{NNConv}\Big( \\
&\quad \text{ReLU}(X W_{\text{FC}} + b_{\text{FC}}), \\
&\quad A, E; W_{\text{NN}}, b_{\text{NN}}\Big), \\
&\quad T; W_{\text{LSTM}}, h\Big), A; W_{\text{GAT}}\Big) \\
\end{align*}

The output of this GCN model is a set of node embeddings, denoted as $\text{Emb}$, where 
\begin{align*}
\text{Emb} &= \text{GCN}(X, A, E, T, P).
\end{align*}

The FAISS L2 index is then used to efficiently search within these embeddings. Let $\text{Query}$ be a query vector of dimension $D$. The FAISS L2 index searches for the nearest neighbors of $\text{Query}$ in the embedding space:
\begin{align*}
\text{NearestNeighbors}(&\text{Query}, \text{Emb}) = \\
&\text{FAISS}_{L2}(\text{Query}, \text{Emb}, D)
\end{align*}

\section{Results}

\noindent We have conducted numerous graph analysis experiments on DAO governance voting with a range of graph sizes, learnable features, specialized convolutional layers, and training and optimization algorithms. We also tested different clustering approaches including DBSCAN and other unsupervised methods; we deployed supervised classification methods with the labelled subgraph of true (Ethereum Name Service) identities; and we considered the problem from different graph analysis viewpoints (such as considering Sybil identification as a problem of link prediction or as a community detection problem). While many of these methods initially looked promising, most failed for a variety of reasons; most commonly, approaches cannot rely on (\emph{k}$>$1) edge connectivity, which is precisely what provides voter anonymity. 

However, our task of identifying Sybils is easier than complete deanonymization and by approaching the problem as a \emph{k}-anonymous graph signal we might achieve what we want anyway, which is the ability to defend against attackers at scale but not reveal individual identities. This is an easier, but not easy problem (avoiding Douceur's \cite{douceur_Sybil_2002} strict definition of a Sybil). Since we should expect Sybils to be singletons, also like voting spam, we cannot effectively cluster their lower dimensional structure, nor can we use a labelled subgraph of known users for supervised learning, and nor will we be able to detect communities or cliques. Instead, we defined Sybils as \emph{clusters of similar higher dimensional vector node embeddings}. And since we form clusters using a \emph{k}-means clustering method on higher dimensional embeddings, singletons pose no particular challenge for our method. 

Graph convolutional neural networks offer several advantages for identifying Sybils. First, network topology is informative but difficult to learn using traditional machine learning methods. This is especially true for very large graphs. However, since convolutional networks aggregate information from nearby nodes, and therefore inductively bias these relationships, learning is simplified (the graph signal is downsampled), and since parameters are shared across convolutional layers (lowering memory requirements), large graphs are less difficult to learn. Moreover, our encoder-decoder approach takes cues from Natural Language Processing (NLP), which must accommodate a large number of token inputs (precluding deep, fully connected networks). Our use of a multi-head graph attention layer produces node embeddings with \emph{n}-deep neighborhood information (this is also known as a transformer), thereby increasing network depth but with tractable sparsity. Similarly, the Long-Short Term Memory (LSTM) algorithm used to learn time series data can be considered an earlier version of the transformer model. More broadly, transformer models, although still rarely applied to graph problems, offer significant untapped potential for network deanonymization and have already proven useful for image processing and other discrete structural problems. 

\begin{figure*}[ht]
    \centering
    \includegraphics[width=\textwidth]{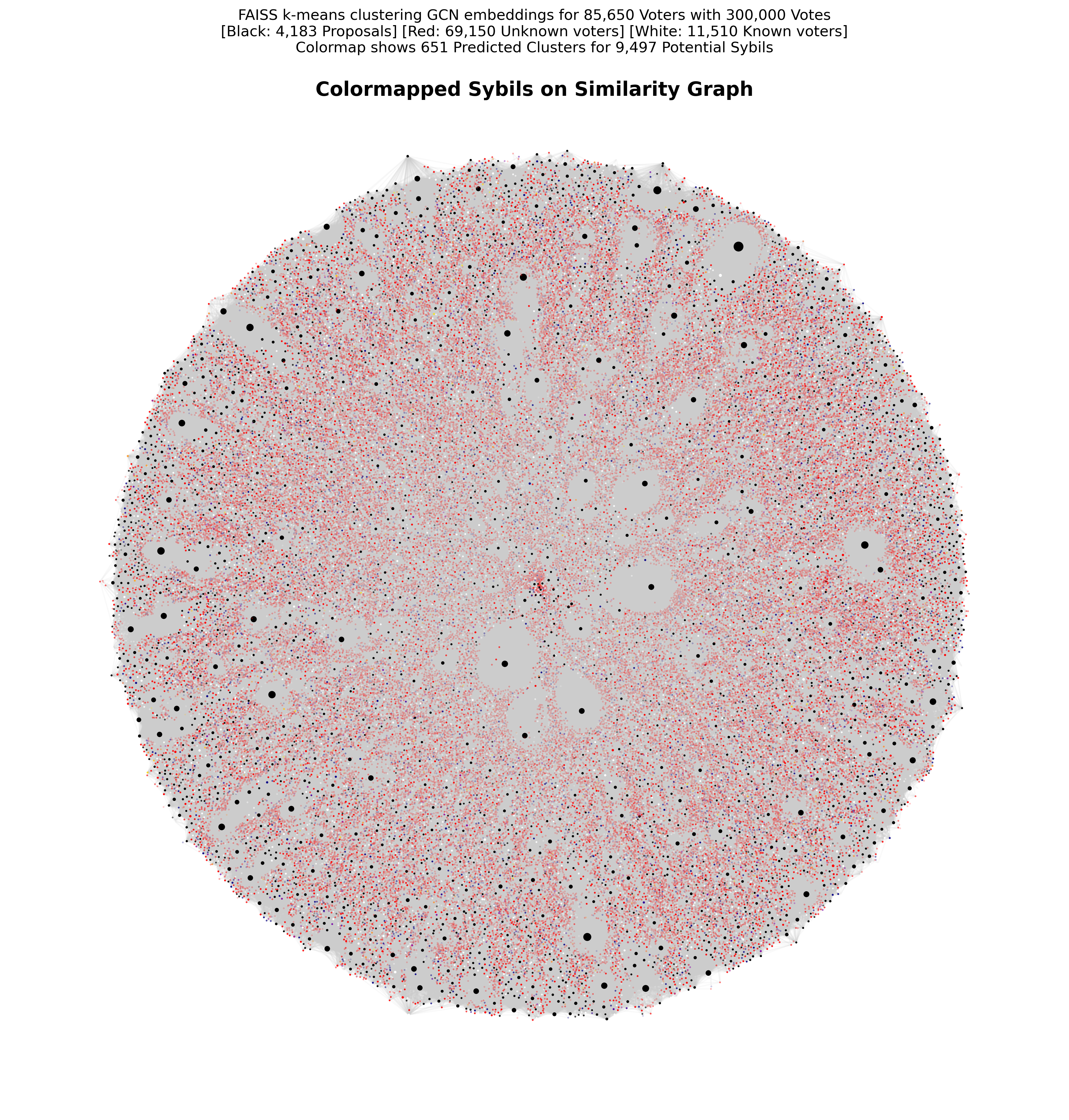}
    \caption{9,497 Potential Sybils Identified in 651 Clusters for 300,000 Votes.}
    \label{fig:similarity_graph}
\end{figure*}

\begin{figure*}[ht]
    \centering
    \includegraphics[width=\textwidth]{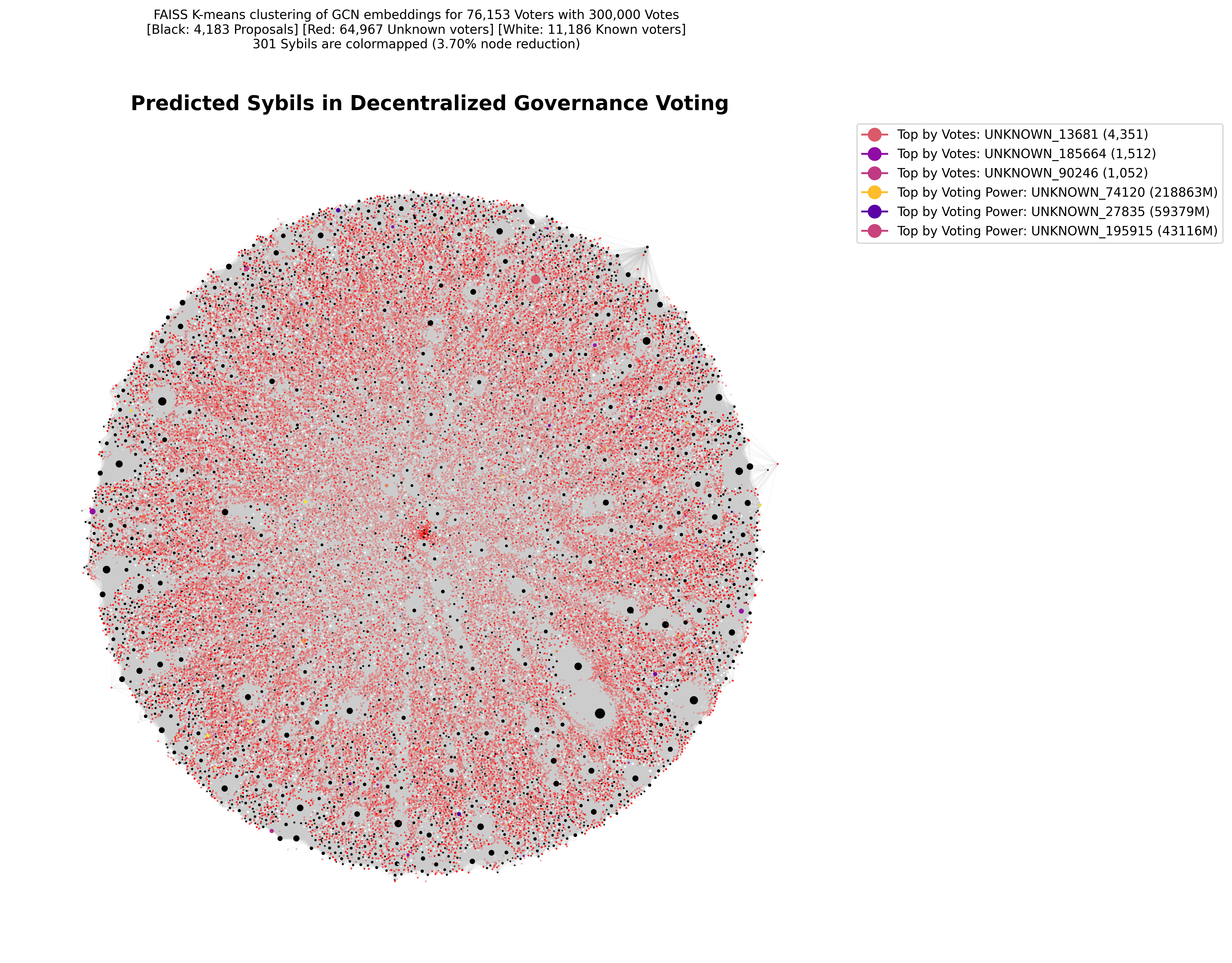}
    \caption{301 Sybils Combined for a 3.7\% Reduction in Voters.}
    \label{fig:clustered_graph}
\end{figure*}

Tested against the snapshot.org dataset, our method typically reduces the original voting graph by approximately 2-5\% (subject to hyperparameters). Quite obviously, performance could be improved and validation measurements need to be established. The method produces a voting graph with real IDs that can be inspected for further forensic analysis or visualized (see Figures \ref{fig:similarity_graph} and \ref{fig:clustered_graph}). Moreover, we can produce a few key sociometric measurements to begin the process of constructing community baselines for further research (see Table \ref{tab:data_analysis}).

\begin{table*}[ht]
\centering
\begin{tabular}{lr}
\hline
\textbf{Metric} & \textbf{Value} \\
\hline
Start Date & 2020-07-18 \\
End Date & 2021-04-20 \\
Duration & 0 years and 275 days \\
Original Graph & 89,833 nodes, 300,000 edges \\
Similarity Graph & 89,833 nodes, 300,000 edges \\
Clustered Graph & 87,874 nodes, 300,000 edges \\
Number of Known Voters & 11,510 \\
Number of Unknown Voters & 69,150 \\
Number of Potential Sybils Identified & 9,947 \\
Number of Sybil Clusters Formed & 301 (0.35\% of Unknown Voters) \\
Node Reduction After Clustering Sybils & 3.7\% \\
\hline
\end{tabular}
\caption{Sociometric Results for 300,000 Votes}
\label{tab:data_analysis}
\end{table*}

\section{Future Research}

\noindent The emergence of Decentralized Autonomous Organizations (DAOs) marks a pivotal shift in the governance of digital assets. DAOs, operating without centralized control, face significant challenges from Sybil attacks, where one entity creates many identities to gain disproportionate influence. Building on Ostrom’s principles for managing commons, we recognize that leniency might not directly apply in the digital realm due to the anonymous nature of offenders. Consequently, once armed with effective tools to identify Sybils, DAOs would still need to experimentally discover governance protocols to mitigate Sybil threats at scale. Penalties must be carefully calibrated to deter malicious behaviour without stifling the collaborative ethos that underpins DAOs.

To improve and further validate our inductive method of Sybil identification we need benchmarks from other online communities. Moreover, more sophisticated and further development of machine learning features would enhance the method. Finally, despite the challenges required to `follow the money,' using this direct method to identify Sybil chains is the most reliable signal we have, and if conducted at scale the method would provide an invaluable graph signal for additional supervised learning. 

We could also use the trained embeddings to generatively produce voting samples, which can be stimulated with varying Sybil signals. By generating a sampled voting signal, we can create an accurate simulation to reflect the temporal dynamics and evolution of online communities. The insight for simulation is that by generating samples from high dimensional embeddings we can represent specific voting behaviours. Further, the introduction of Sybil signals reveals potential sources of resource congestion and over-extraction.

In addition to refining current methods, it is worth exploring other deep learning techniques. To adapt the method for detecting voting spam, we could instead detect anomalies within the embeddings. Instead of identifying similar nodes, searching for outliers might indicate spam activity (perhaps this might be exploited using transfer learning). Or, since the GCNN trains an embeddings model, generative tasks like link prediction or more complex temporal-economic simulation might be possible.

Finally, machine learning methods can be exploited by malicious users. Making machine learning methods resilient to attack is an active area of research and a practical requirement for DAOs, which must protect their digital commons from theft and inappropriate extraction. Examples of attacks against ML models include adversarial attacks that manipulate input data to deceive models, model inversion attacks that reveal sensitive information, and poisoning attacks that corrupt training data. Developing robust models that can detect and resist these attacks is crucial and should be part of any production environment (see \cite{buterin_promise_2024}). Additionally, understanding and mitigating the ethical implications of AI and ML in DAO governance, such as bias in decision-making processes, must be a priority. These challenges open up diverse research avenues, ranging from developing new algorithms for anomaly detection to ethical frameworks for AI deployment in decentralized governance structures.

\section{Conclusion}
The Internet’s transformative impact on communication costs has laid the groundwork for a similar paradigm shift in governance technologies. By drastically reducing the costs associated with dispute resolution and collaboration, these technologies have the potential to alter the landscape of online life. This shift is frustrated by anonymous accounts called Sybils, but by deploying a deep learning method we augmented network transparency to minimize the costs of monitoring and enforcement within governance structures. Without impacting personal privacy and without relying on unproven identity systems, our approach modulates governance behaviours at scale to prevent Sybil attacks, potentially heralding a new era of efficiency and effectiveness in governance systems.

The concept of Digital Common Pool Resources (DCPRs) and polycentric governance emerges as a promising avenue for future research in this context. This governance model is particularly relevant in addressing the perennial challenges associated with managing digital assets in a commons, such as congestion, overuse, pollution, and theft. Traditional approaches often struggle with enforcing limits on resource utilization, but polycentric governance offers a novel framework that could be more effective in managing these resources. With polycentric governance, participants in the digital commons develop context-specific rules to prevent infringement and establish consequences for those who exploit the commons. By leveraging decentralized decision-making and consensus mechanisms, these regimes hold the potential for more sustainable and equitable digital resource management.

The emerging field of DAO Studies provides insightful parallels for polycentric governance. DAOs operate on principles of distributed authority and decision-making, which align closely with the tenets of polycentric governance. This alignment offers a unique opportunity to draw insights from DAOs to inform traditional policy, regulation, and governmental contexts. The lessons learned from DAO governance models, such as consensus building, decentralized control, and community-driven decision-making, could offer valuable strategies for tackling governance challenges in a broader context. By integrating these insights, traditional governance systems could evolve to become more adaptive, participatory, and responsive to the needs of their constituencies.

\bibliographystyle{plain}
\bibliography{main}

\begin{thebibliography}{10}

\bibitem{akcora_bitcoinheist_2019}
Cuneyt~Gurcan Akcora, Yitao Li, Yulia~R. Gel, and Murat Kantarcioglu.
\newblock {BitcoinHeist}: {Topological} {Data} {Analysis} for {Ransomware} {Detection} on the {Bitcoin} {Blockchain}, June 2019.
\newblock arXiv:1906.07852 [cs].

\bibitem{ashfaq_machine_2022}
Tehreem Ashfaq, Rabiya Khalid, Adamu~Sani Yahaya, Sheraz Aslam, Ahmad~Taher Azar, Safa Alsafari, and Ibrahim~A. Hameed.
\newblock A {Machine} {Learning} and {Blockchain} {Based} {Efficient} {Fraud} {Detection} {Mechanism}.
\newblock {\em Sensors}, 22(19):7162, January 2022.

\bibitem{axelsen_when_2022}
Henrik Axelsen, Johannes~Rude Jensen, and Omri Ross.
\newblock When is a {DAO} {Decentralized}?
\newblock {\em Complex Systems Informatics and Modeling Quarterly}, (31):51--75, July 2022.

\bibitem{backstrom_wherefore_2007}
Lars Backstrom, Cynthia Dwork, and Jon Kleinberg.
\newblock Wherefore {Art} {Thou} {R3579x}? {Anonymized} {Social} {Networks}, {Hidden} {Patterns}, and {Structural} {Steganography}.
\newblock In {\em Proceedings of the 16th international conference on {World} {Wide} {Web}}, {WWW} '07, pages 181--190, New York, NY, USA, May 2007. Association for Computing Machinery.

\bibitem{bellavitis_rise_2023}
Cristiano Bellavitis, Christian Fisch, and Paul~P. Momtaz.
\newblock The rise of decentralized autonomous organizations ({DAOs}): a first empirical glimpse.
\newblock {\em Venture Capital}, 25(2):187--203, April 2023.
\newblock Publisher: Routledge \_eprint: https://doi.org/10.1080/13691066.2022.2116797.

\bibitem{benamar_identification_2017}
Lamya Benamar, Christine Balagué, and Mohamad Ghassany.
\newblock The {Identification} and {Influence} of {Social} {Roles} in a {Social} {Media} {Product} {Community}.
\newblock {\em Journal of Computer-Mediated Communication}, 22(6):337--362, November 2017.

\bibitem{bitmex_research_bitcoin_2022}
{BitMEX Research}.
\newblock Bitcoin {Address} {Re}-use {Statistics}, February 2022.

\bibitem{blackburn_cooperation_2022}
Alyssa Blackburn, Christoph Huber, Yossi Eliaz, Muhammad~S. Shamim, David Weisz, Goutham Seshadri, Kevin Kim, Shengqi Hang, and Erez~Lieberman Aiden.
\newblock Cooperation among an anonymous group protected {Bitcoin} during failures of decentralization, June 2022.

\bibitem{braun_collusion-proof_2022}
Alexander Braun, Niklas Häusle, and Stephan Karpischek.
\newblock Collusion-{Proof} {Decentralized} {Autonomous} {Organizations}, April 2022.

\bibitem{buterin_promise_2024}
Vitalik Buterin.
\newblock The promise and challenges of crypto + {AI} applications, January 2024.

\bibitem{carillo_what_2017}
Kevin Carillo, Sid Huff, and Brenda Chawner.
\newblock What makes a good contributor? {Understanding} contributor behavior within large {Free}/{Open} {Source} {Software} projects – {A} socialization perspective.
\newblock {\em The Journal of Strategic Information Systems}, 26(4):322--359, December 2017.

\bibitem{dahiya_neural_2023}
Manju Dahiya, Naman Mishra, Riya Singh, and {Pavitra}.
\newblock Neural network based approach for {Ethereum} fraud detection.
\newblock In {\em 2023 4th {International} {Conference} on {Intelligent} {Engineering} and {Management} ({ICIEM})}, pages 1--4, London, United Kingdom, May 2023. IEEE.

\bibitem{dahlander_progressing_2011}
Linus Dahlander and Siobhan O'Mahony.
\newblock Progressing to the {Center}: {Coordinating} {Project} {Work}.
\newblock {\em Organization Science}, 22(4):961--979, August 2011.

\bibitem{dejean_big_2015}
Sylvain Dejean and Nicolas Jullien.
\newblock Big from the beginning: {Assessing} online contributors’ behavior by their first contribution.
\newblock {\em Research Policy}, 44(6):1226--1239, July 2015.

\bibitem{douceur_Sybil_2002}
John~R. Douceur.
\newblock The {Sybil} {Attack}.
\newblock International Workshop on Peer-to-Peer Systems, 2002.

\bibitem{dupont_experiments_2018}
Quinn DuPont.
\newblock Experiments in {Algorithmic} {Governance}: {An} ethnography of "{The} {DAO}," a failed {Decentralized} {Autonomous} {Organization}.
\newblock In Malcolm Campbell-Verduyn, editor, {\em Bitcoin and {Beyond}: {The} {Challenges} and {Opportunities} of {Blockchains} for {Global} {Governance}}, pages 157--177. Routledge, New York, 2018.

\bibitem{dupont_guiding_2020}
Quinn DuPont.
\newblock Guiding {Principles} for {Ethical} {Cryptocurrency}, {Blockchain}, and {DLT} {Research}.
\newblock {\em Cryptoeconomic Systems}, 1(1), 2020.

\bibitem{dupont_progressive_2024}
Quinn DuPont.
\newblock A {Progressive} {Web3}: {From} {Digital} {Polycentric} {Governance} to {Social} {Coproduction}.
\newblock In {\em Defining {Web3}: {A} {Guide} to the {New} {Cultural} {Economy}}, Research in the {Sociology} of {Organizations}. Emerald Group Publishing, 2024.

\bibitem{dylan-ennis_hash_2023}
Paul Dylan-Ennis.
\newblock Hash, {Bash}, {Cash}: {How} {Changed} {Happens} in {Decentralised} {Web3} {Cultures}.
\newblock 2023.

\bibitem{fan_altruistic_2023}
Sizheng Fan, Tian Min, Xiao Wu, and Wei Cai.
\newblock Altruistic and {Profit}-oriented: {Making} {Sense} of {Roles} in {Web3} {Community} from {Airdrop} {Perspective}.
\newblock In {\em Proceedings of the 2023 {CHI} {Conference} on {Human} {Factors} in {Computing} {Systems}}, {CHI} '23, pages 1--16, New York, NY, USA, April 2023. Association for Computing Machinery.

\bibitem{faqir-rhazoui_comparative_2021}
Youssef Faqir-Rhazoui, Javier~Arroyo Gallardo, and Samer Hassan.
\newblock A {Comparative} {Analysis} of the {Adoption} of {Decentralized} {Governance} in the {Blockchain} {Through} {DAOs}.
\newblock preprint, In Review, February 2021.

\bibitem{mcgill_university_leading_2015}
Samer Faraj, Srinivas Kudaravalli, and Molly Wasko.
\newblock Leading {Collaboration} in {Online} {Communities}.
\newblock {\em MIS Quarterly}, 39(2):393--412, February 2015.

\bibitem{faraj_online_2016}
Samer Faraj, Georg von Krogh, Eric Monteiro, and Karim~R. Lakhani.
\newblock Online {Community} as {Space} for {Knowledge} {Flows}.
\newblock {\em Information Systems Research}, 27(4):668--684, December 2016.

\bibitem{feeny_tragedy_1990}
David Feeny, Fikret Berkes, Bonnie~J. McCay, and James~M. Acheson.
\newblock The {Tragedy} of the {Commons}: {Twenty}-{Two} {Years} {Later}.
\newblock {\em Human Ecology}, 18(1):1--19, 1990.

\bibitem{filippi_blockchain_2016}
Primavera~De Filippi and Samer Hassan.
\newblock Blockchain technology as a regulatory technology: {From} code is law to law is code.
\newblock {\em First Monday}, 21(12), November 2016.

\bibitem{gallus_fostering_2017}
Jana Gallus.
\newblock Fostering {Public} {Good} {Contributions} with {Symbolic} {Awards}: {A} {Large}-{Scale} {Natural} {Field} {Experiment} at {Wikipedia}.
\newblock {\em Management Science}, 63(12):3999--4015, December 2017.

\bibitem{goldberg_metaverse_2023}
Mitchell Goldberg and Fabian Schär.
\newblock Metaverse governance: {An} empirical analysis of voting within {Decentralized} {Autonomous} {Organizations}.
\newblock {\em Journal of Business Research}, 160:113764, May 2023.

\bibitem{halaburda_digitization_2023}
Hanna Halaburda, Natalia Levina, and Min Semi.
\newblock Digitization of {Transaction} {Terms} as a {Shift} {Parameter} within {TCE}: {Strong} {Smart} {Contract} as a {New} {Mode} of {Transaction} {Governance}, June 2023.

\bibitem{hardin_tragedy_1968}
Garrett Hardin.
\newblock The {Tragedy} of the {Commons}.
\newblock {\em Science}, 162(3859):1243--1248, 1968.

\bibitem{hardin_tragedy_1994}
Garrett Hardin.
\newblock The tragedy of the unmanaged commons.
\newblock {\em Trends in Ecology \& Evolution}, 9(5):199, May 1994.

\bibitem{hirschman_exit_1970}
Albert~O. Hirschman.
\newblock {\em Exit, {Voice}, and {Loyalty}: {Responses} to {Decline} in {Firms}, {Organizations}, and {States}}.
\newblock Harvard University Press, Cambridge MA, 1970.

\bibitem{jain_plural_2023}
Shrey Jain, Divya Siddarth, and Glen Weyl.
\newblock Plural {Publics}.
\newblock March 2023.

\bibitem{kappos_how_2022}
George Kappos, Haaroon Yousaf, Rainer Stütz, Soﬁa Rollet, and Bernhard Haslhofer.
\newblock How to {Peel} a {Million}: {Validating} and {Expanding} {Bitcoin} {Clusters}.
\newblock 2022.

\bibitem{lemieux_searching_2022}
Victoria Lemieux.
\newblock {\em Searching for {Trust}: {Blockchain} {Technology} in an {Age} of {Disinformation}}.
\newblock Cambridge University Press, Cambridge UK, 2022.

\bibitem{liu_illusion_2023}
Xuan Liu.
\newblock The {Illusion} of {Democracy}? {An} {Empirical} {Study} of {DAO} {Governance} and {Voting} {Behavior}, May 2023.

\bibitem{lorenz_machine_2021}
Joana Lorenz, Maria~Inês Silva, David Aparício, João~Tiago Ascensão, and Pedro Bizarro.
\newblock Machine learning methods to detect money laundering in the bitcoin blockchain in the presence of label scarcity.
\newblock In {\em Proceedings of the {First} {ACM} {International} {Conference} on {AI} in {Finance}}, {ICAIF} '20, pages 1--8, New York, NY, USA, October 2021. Association for Computing Machinery.

\bibitem{mcinnes_umap_2020}
Leland McInnes, John Healy, and James Melville.
\newblock {UMAP}: {Uniform} {Manifold} {Approximation} and {Projection} for {Dimension} {Reduction}, September 2020.
\newblock arXiv:1802.03426 [cs, stat].

\bibitem{meiklejohn_fistful_2013}
Sarah Meiklejohn, Marjori Pomarole, Grant Jordan, Kirill Levchenko, Damon McCoy, Geoffrey~M. Voelker, and Stefan Savage.
\newblock A {Fistful} of {Bitcoins}: {Characterizing} {Payments} {Among} {Men} with {No} {Names}.
\newblock In {\em Proceedings of the 2013 {Conference} on {Internet} {Measurement} {Conference}}, {IMC} '13, pages 127--140, New York, NY, USA, 2013. ACM.

\bibitem{miller_beyond_2022}
Joel Miller, Eric~Glen Weyl, and Leon Erichsen.
\newblock Beyond {Collusion} {Resistance}: {Leveraging} {Social} {Information} for {Plural} {Funding} and {Voting}.
\newblock {\em SSRN Electronic Journal}, 2022.

\bibitem{mindel_sustainability_2018}
Vitali Mindel, Lars Mathiassen, and Arun Rai.
\newblock The {Sustainability} of {Polycentric} {Information} {Commons}.
\newblock {\em MIS Quarterly}, 42(2):607--631, February 2018.

\bibitem{moser_resurrecting_2022}
Malte Möser and Arvind Narayanan.
\newblock Resurrecting {Address} {Clustering} in {Bitcoin}, July 2022.

\bibitem{narayanan_-anonymizing_2009}
Arvind Narayanan and Vitaly Shmatikov.
\newblock De-anonymizing {Social} {Networks}.
\newblock In {\em 2009 30th {IEEE} {Symposium} on {Security} and {Privacy}}, pages 173--187, May 2009.

\bibitem{ostrom_governing_1990}
Elinor Ostrom.
\newblock {\em Governing the {Commons}}.
\newblock Cambridge University Press, Cambridge UK, 1990.

\bibitem{ostrom_revisiting_1999}
Elinor Ostrom, Joanna Burger, Christopher~B. Field, Richard~B. Norgaard, and David Policansky.
\newblock Revisiting the {Commons}: {Local} {Lessons}, {Global} {Challenges}.
\newblock {\em Science}, 284(5412):278--282, April 1999.

\bibitem{owocki_33_2022}
Kevin Owocki and Bryan Ford.
\newblock Sybil resistance with bryan ford.

\bibitem{paquet-clouston_ransomware_2018}
Masarah Paquet-Clouston, Bernhard Haslhofer, and Benoit Dupont.
\newblock Ransomware {Payments} in the {Bitcoin} {Ecosystem}, April 2018.
\newblock arXiv:1804.04080 [cs].

\bibitem{pham_anomaly_2017}
Thai Pham and Steven Lee.
\newblock Anomaly {Detection} in {Bitcoin} {Network} {Using} {Unsupervised} {Learning} {Methods}, February 2017.
\newblock arXiv:1611.03941 [cs].

\bibitem{prince_understanding_2024}
Simon~J.D. Prince.
\newblock {\em Understanding {Deep} {Learning}}.
\newblock MIT Press, 2024.

\bibitem{rabieinejad_generative_2023}
Elnaz Rabieinejad, Abbas Yazdinejad, Reza~M. Parizi, and Ali Dehghantanha.
\newblock Generative {Adversarial} {Networks} for {Cyber} {Threat} {Hunting} in {Ethereum} {Blockchain}.
\newblock {\em Distributed Ledger Technologies: Research and Practice}, 2(2):9:1--9:19, June 2023.

\bibitem{raymond_cathedral_1999}
Eric~S. Raymond.
\newblock {\em The {Cathedral} \& the {Bazaar}: {Musings} on {Linux} and {Open} {Source} by an {Accidental} {Revolutionary}}.
\newblock O'Reilly, Cambridge MA, 1st ed edition, 1999.

\bibitem{rikken_ins_2021}
Olivier Rikken, Marijn Janssen, and Zenlin Kwee.
\newblock The {Ins} and {Outs} of {Decentralized} {Autonomous} {Organizations} ({Daos}), December 2021.

\bibitem{sankar_roy_exploiting_2022}
Kowshik Sankar~Roy, Md. Ebtidaul~Karim, and Pritom Biswas~Udas.
\newblock Exploiting {Deep} {Learning} {Based} {Classification} {Model} for {Detecting} {Fraudulent} {Schemes} over {Ethereum} {Blockchain}.
\newblock In {\em 2022 4th {International} {Conference} on {Sustainable} {Technologies} for {Industry} 4.0 ({STI})}, pages 1--6, December 2022.

\bibitem{schneider_exit_2021}
Nathan Schneider and Morshed Mannan.
\newblock Exit to {Community}: {Strategies} for {Multi}-{Stakeholder} {Ownership} in the {Platform} {Economy}.
\newblock {\em Georgetown Law Technology Review}, 5(1):71, 2021.

\bibitem{sharma_unpacking_2023}
Tanusree Sharma, Yujin Kwon, Kornrapat Pongmala, Henry Wang, Andrew Miller, Dawn Song, and Yang Wang.
\newblock Unpacking {How} {Decentralized} {Autonomous} {Organizations} ({DAOs}) {Work} in {Practice}, April 2023.

\bibitem{sun_voter_2023}
Xiaotong Sun, Xi~Chen, Charalampos Stasinakis, and Georgios Sermpinis.
\newblock Voter {Coalitions} in {Decentralized} {Autonomous} {Organization} ({DAO}): {Evidence} from {MakerDAO}, January 2023.

\bibitem{sun_decentralization_2023}
Xiaotong Sun, Charalampos Stasinakis, and Georigios Sermpinis.
\newblock Decentralization illusion in {Decentralized} {Finance}: {Evidence} from tokenized voting in {MakerDAO} polls, March 2023.

\bibitem{sweeney_k-anonymity_2002}
Latanya Sweeney.
\newblock K-{Anonymity}: {A} {Model} for {Protecting} {Privacy}.
\newblock {\em International Journal of Uncertainty, Fuzziness and Knowledge-Based Systems}, 10(05):557--570, October 2002.

\bibitem{vergne_web3_2023}
JP~Vergne.
\newblock Web3 as {Decentralization} {Theater}? {A} {Framework} for {Envisioning} {Decentralization} {Strategically}.
\newblock 2023.

\bibitem{wang_empirical_2023}
Qin Wang, Guangsheng Yu, Yilin Sai, Caijun Sun, Lam~Duc Nguyen, Sherry Xu, and Shiping Chen.
\newblock An {Empirical} {Study} on {Snapshot} {DAOs}, May 2023.

\bibitem{weber_anti-money_2019}
Mark Weber, Giacomo Domeniconi, Jie Chen, Daniel Karl~I. Weidele, Claudio Bellei, Tom Robinson, and Charles~E. Leiserson.
\newblock Anti-{Money} {Laundering} in {Bitcoin}: {Experimenting} with {Graph} {Convolutional} {Networks} for {Financial} {Forensics}, July 2019.
\newblock arXiv:1908.02591 [cs, q-fin].

\bibitem{williamson_transaction-cost_1979}
Oliver~E. Williamson.
\newblock Transaction-{Cost} {Economics}: {The} {Governance} of {Contractual} {Relations}.
\newblock {\em The Journal of Law \& Economics}, 22(2):233--261, 1979.

\bibitem{wright_measuring_2021}
Steven~A. Wright.
\newblock Measuring {DAO} {Autonomy}: {Lessons} {From} {Other} {Autonomous} {Systems}.
\newblock {\em IEEE Transactions on Technology and Society}, 2(1):43--53, 2021.

\bibitem{wu_detecting_2022}
Jiajing Wu, Jieli Liu, Weili Chen, Huawei Huang, Zibin Zheng, and Yan Zhang.
\newblock Detecting {Mixing} {Services} via {Mining} {Bitcoin} {Transaction} {Network} with {Hybrid} {Motifs}.
\newblock {\em IEEE Transactions on Systems, Man, and Cybernetics: Systems}, 52(4):2237--2249, April 2022.
\newblock arXiv:2001.05233 [cs].

\bibitem{xu_autogov_2023}
Jiahua Xu, Daniel Perez, Yebo Feng, and Benjamin Livshits.
\newblock Auto.gov: {Learning}-based {On}-chain {Governance} for {Decentralized} {Finance} ({DeFi}), February 2023.

\bibitem{zhao_task_2022}
Xi~Zhao, Peilin Ai, Fujun Lai, Xin~(Robert) Luo, and Jose Benitez.
\newblock Task management in decentralized autonomous organization.
\newblock {\em Journal of Operations Management}, 68(6-7):649--674, 2022.

\bibitem{zhou_behavior-aware_2022}
Jiajun Zhou, Chenkai Hu, Jianlei Chi, Jiajing Wu, Meng Shen, and Qi~Xuan.
\newblock Behavior-aware {Account} {De}-anonymization on {Ethereum} {Interaction} {Graph}.
\newblock {\em IEEE Transactions on Information Forensics and Security}, 17:3433--3448, 2022.
\newblock arXiv:2203.09360 [cs].

\bibitem{ziegler_taxonomy_2022}
Christian Ziegler and Isabell Welpe.
\newblock A {Taxonomy} of {Decentralized} {Autonomous} {Organizations}.
\newblock {\em ICIS 2022 Proceedings}, December 2022.

\end{thebibliography}

\newpage
\appendix
\section*{Appendix: Retrieval Augmented Generation (RAG) of Software Engineering Issues for Monero Cryptocurrency: Hierarchical GPT-4 Turbo for Automated Histories*}\label{appendix}

\noindent * This text was automatically generated.\newline

\noindent Monero, a cryptocurrency project prioritizing privacy and security, has traversed a complex journey marked by technical innovations, governance shifts, and community dynamics. This report meticulously chronicles the pivotal events, challenges, and actors within Monero's evolution, providing a forensic analysis of its development, governance, and community engagement.

\subsection*{Technical Developments and Challenges}

\begin{enumerate}
    \item \textbf{Kovri Development and Windows Sessions Issue (2017-2018):}
    The Kovri project, aimed at enhancing Monero's privacy through IP anonymization, faced significant hurdles in its Windows implementation. Developers \texttt{@danrmiller}, \texttt{@Jaqueeee}, \texttt{@anonimal}, and \texttt{@luigi1111} navigated logout issues on Remote Desktop Protocol (RDP) disconnects, spotlighting the technical complexities of privacy-focused software development. Despite the resolution, this episode underscored vulnerabilities in remote development practices and hinted at the intricate balance between security features and user experience.
    
    \item \textbf{BuildBot Efficiency and Documentation-only Changes (2019):}
    Spearheaded by \texttt{@danrmiller}, this initiative addressed inefficiencies in Monero's automated build system, where minor documentation changes unnecessarily consumed computational resources. The issue's formal ticketing illustrated a proactive approach to optimizing development operations but also exposed potential bottlenecks in Monero's resource allocation strategies.
    
    \item \textbf{Seraphis Wallet Workgroup Meeting Discrepancies (2020-2021):}
    The Seraphis Wallet Workgroup, with contributors like \texttt{ghostway}, \texttt{rbrunner7}, \texttt{sneedlewoods\_xmr}, and \texttt{jberman}, encountered coordination challenges and technical obstacles in wallet development. Despite these hurdles, the group's regular meetings and collaborative resolve showcased Monero's community-driven approach to innovation.
\end{enumerate}

\subsection*{Governance and Oversight}

\begin{enumerate}
    \setcounter{enumi}{3}
    \item \textbf{RandomX Evaluation and Adjustment Proposals (2022):}
    With evolving hardware technologies, the Monero community debated the future of its Proof-of-Work algorithm, RandomX. This discourse, led by community contributors, highlighted the perennial challenge of maintaining an equitable mining landscape amidst rapid technological advancements, emphasizing the need for adaptive algorithmic responses.
    
    \item \textbf{Moderation Discrepancies in Monero Channels (2023):}
    Issues with moderation practices within Monero's communication channels surfaced, revealing inconsistencies in banning decisions and handling of threats. This episode pointed to the broader challenge of maintaining open yet respectful dialogues in decentralized communities.
\end{enumerate}

\subsection*{Community Engagement and Technical Support}

\begin{enumerate}
    \setcounter{enumi}{5}
    \item \textbf{CCS Escrow Concerns (2024):}
    Following incidents of theft, the Monero community questioned the continuation of Luigi as CCS escrow, leading to a vote. This event not only highlighted security vulnerabilities within the project's crowdfunding mechanism but also sparked discussions on enhancing trust and transparency in community-driven funding.
    
    \item \textbf{Debate on Deprecating hardware@getmonero.org Email (2024):}
    Proposals for public audits and a public directory emerged amid concerns over the transparency and potential misuse of the official email. This debate reflected broader issues of operational transparency and governance within the Monero project.
\end{enumerate}

\subsection*{Monero Governance Summary}

Throughout its history, Monero has faced numerous technical, governance, and community engagement challenges. From addressing technical intricacies in its privacy mechanisms to navigating governance reforms and fostering community trust, Monero's journey is emblematic of the ongoing struggle to balance privacy, security, and decentralization in cryptocurrency development. The project's resilience and adaptability, underscored by its community's commitment to privacy and security, remain its defining features. As Monero continues to evolve, this forensic history underscores the critical importance of vigilance, adaptability, and community consensus in safeguarding its future.

\end{document}